%% file: main.tex
\documentclass{article} 
\usepackage{iclr2026_conference,times}

\input{math_commands.tex}

\usepackage[table]{xcolor}
\usepackage[dvipsnames, svgnames, x11names]{xcolor}

\usepackage{hyperref}
\usepackage{url}
\let\eqref\relax
\DeclareRobustCommand{\eqref}[1]{\textup{{Eq. (\ref{#1})}}}
\usepackage[utf8]{inputenc} 
\usepackage[T1]{fontenc}    
\hypersetup{hidelinks,
    colorlinks=true,
    anchorcolor=blue,
    citecolor={olive},
    linkcolor=blue}
\usepackage{booktabs}       
\usepackage{amsfonts}       
\usepackage{nicefrac}       
\usepackage{microtype}      
\usepackage{xcolor}         
\usepackage{times}
\usepackage{soul}
\usepackage{graphicx}
\usepackage{amsmath}
\usepackage{amsthm}
\usepackage{amssymb}
\usepackage{booktabs}
\usepackage{algorithm}
\usepackage{algpseudocode}
\usepackage{subfig}
\usepackage{chngpage}
\usepackage{wrapfig}
\usepackage{enumitem}
\usepackage{multirow}
\usepackage{subfloat}
\usepackage{color}  

\newtheorem{lemma}{Lemma}

\newcommand{\Exp}{\mathop{\mathbb E}\displaylimits}

\usepackage{booktabs}
\usepackage{siunitx}

\title{Mind Your Entropy: From Maximum Entropy to Trajectory Entropy-Constrained RL
}

\author{Guojian Zhan$^{1,2}$, Likun Wang$^{1}$, Pengcheng Wang$^{2}$, Feihong Zhang$^{1}$,  \\ \textbf{Jingliang Duan}$^{3}$, \textbf{Masayoshi Tomizuka}$^{2}$, \textbf{Shengbo Eben Li}$^{1}$\footnotemark[1]\\
$^1$ School of Vehicle and Mobility, Tsinghua University\\
$^2$ Department of Mechanical Engineering, UC Berkeley\\
$^3$ School of Mechanical Engineering, University of Science and Technology Beijing\\
\texttt{zgj21@mails.tsinghua.edu.cn,  lishbo@tsinghua.edu.cn}\\
}

\iclrfinalcopy 
\begin{document}

\maketitle

\begin{abstract}
Maximum entropy has become a mainstream off-policy reinforcement learning (RL) framework for balancing exploitation and exploration. However, two bottlenecks still limit further performance improvement: \textit{(1) non-stationary Q-value estimation} caused by jointly injecting entropy and updating its weighting parameter, i.e., temperature; and \textit{(2) short-sighted local entropy tuning} that adjusts temperature only according to the current single-step entropy, without considering the effect of cumulative entropy over time. In this paper, we extends maximum entropy framework by proposing a trajectory entropy-constrained reinforcement learning (TECRL) framework to address these two challenges. Within this framework, we first separately learn two Q-functions, one associated with reward and the other with entropy, ensuring clean and stable value targets unaffected by temperature updates. Then, the dedicated entropy Q-function,  explicitly quantifying the expected cumulative entropy, enables us to enforce a trajectory entropy constraint and consequently control the policy’s long-term stochasticity. Building on this TECRL framework, we develop a practical off-policy algorithm, DSAC-E, by extending the state-of-the-art distributional soft actor-critic with three refinements (DSAC-T). Empirical results on the OpenAI Gym benchmark demonstrate that our DSAC-E can achieve higher returns and better stability.
\end{abstract}

\section{Introduction}

Balancing exploration and exploitation remains a central challenge in reinforcement learning (RL)~\citep{sutton2018reinforcement, li2023rlbook}. To address this, off-policy methods have leveraged the maximum entropy principle, which encourages agents to act as randomly as possible while still optimizing for high returns~\citep{wang2022deep, haarnoja2017reinforcement}. By augmenting the objective with a temperature-weighted entropy term, algorithms like Soft Actor-Critic (SAC) \citep{Haarnoja2018SAC} and its distributional variant DSAC \citep{duan2021distributional, duan2025distributional} have achieved state-of-the-art performance on continuous control benchmarks like MuJoCo, proving to be highly effective and robust~\citep{eysenbachmaximum}.

However, a fixed temperature parameter can lead to a policy that is either excessively stochastic or unnecessarily deterministic~\citep{rawlik2012stochastic}. This is because a single temperature value cannot optimally balance exploration and exploitation across all phases of training; a high temperature may hinder convergence, while a low temperature can lead to premature exploitation of a suboptimal solution~\citep{fox2016taming}. To mitigate this issue, modern maximum entropy RL incorporates an automated temperature adjustment mechanism~\citep{Haarnoja2018ASAC}. Using the policy’s current per-step entropy as a feedback signal, this mechanism dynamically tunes the temperature throughout training, aligning it with a predefined target. Therefore, it ensures that a desired level of stochasticity is maintained across all  situations~\citep{hazan2019provably}.

Despite the remarkable empirical success, maximum entropy methods still face two critical bottlenecks that hinder further progress. 
(1) The first issue is \textit{non-stationary Q-value estimation}, which stems from the tight coupling of reward and entropy~\citep{schulman2017PG_Soft-Q}. Since the temperature parameter is updated simultaneously, the injected temperature-weighted entropy term is directly altering the Q-value targets, causing them to become non-stationary.
This process can destabilize value learning and ultimately undermine policy optimization~\citep{lillDDPGicrap2015DDPG}.
(2) Second, and perhaps more fundamentally, while some works have explored constraining entropy, they all suffer from \textit{short-sighted local entropy tuning}~\citep{Haarnoja2018ASAC, duan2021distributional, duan2025distributional}. By regulating only the local current-step entropy, these methods neglect the long-term influence of stochasticity over entire trajectories. More critically, they enforce a uniform entropy target across all states, as if every situation demands the same degree of randomness. This one-size-fits-all assumption is overly restrictive; it fails to acknowledge that effective exploration should adapt to the underlying dynamics and the agent's learning progress~\citep{tokic2010adaptive, sun2022exploit}. This fundamental disconnect ignores the varying exploration needs of different situations.

The observed two bottlenecks naturally raise a question: \textit{can we move beyond maximum entropy by directly and cleanly controlling what really matters—the cumulative entropy of the policy?} We argue the answer is yes by introducing a trajectory entropy-constrained (TEC) RL framework. To ensure a stable and interpretable learning process, our core innovation is to completely decouple the reward and entropy signals by learning two separate Q-functions. This separation ensures clean Q-value targets, and the dedicated entropy critic enables us to enforce a trajectory-level constraint on the policy's cumulative entropy. This design inherently breaks from traditional single-step restriction, enabling a more principled and long-term control of policy stochasticity.


To demonstrate the practical advantages of our framework, we introduce DSAC-E, an extension of the state-of-the-art Distributional Soft Actor-Critic with Three refinements (DSAC-T) algorithm~\citep{duan2025distributional}. DSAC-E integrates the strengths of DSAC-T’s distributional value estimation with our proposed trajectory entropy constraint. By decoupling the reward and entropy Q-values and adjusting the trajectory-level entropy budget, our DSAC-E achieves cleaner and more effective exploitation alongside more controllable exploration. Empirical results on the OpenAI Gym continuous control benchmark~\citep{brockman2016openaigym} demonstrate that DSAC-E not only achieves superior final returns but also exhibits better training stability than strong maximum entropy baselines.

Our contributions are summarized in threefold:
\begin{itemize}
    \item We identify and analyze the impact of two bottlenecks in conventional maximum entropy RL: \textit{(1) non-stationary Q-value estimation} and \textit{(2) short-sighted local entropy tuning}. These issues motivate us to execute reward-entropy separation and trajectory-level entropy constraint;
    \item To address these two identified bottlenecks, we propose the TECRL framework. Within this framework, we first eliminate the \textit{(1) non-stationary Q-value estimation} problem by decoupling reward and entropy signals into two separate critics, while temperature is excluded from the learning processes of both critics. Then the dedicated entropy critic allows us to enforce a trajectory-level entropy constraint, thereby overcoming the issue of \textit{(2) short-sighted local entropy tuning}. Furthermore, we provide a rigorous theoretical analysis demonstrating that appropriately selecting a trajectory entropy budget can yield a higher performance bound;
    \item We introduce DSAC-E, a practical instantiation of our TECRL framework built on DSAC-T, the state-of-the-art maximum entropy algorithm. Through this instantiation, we demonstrate that our framework enables superior performance on complex continuous control tasks.
\end{itemize}


\section{Preliminaries}
\label{sec_preliminaries}



\paragraph{Maximum entropy RL.}
\label{sec:max_entropy}
While standard RL seeks a policy that maximizes the expected accumulated return, maximum entropy RL ~\citep{ haarnoja2017reinforcement} extends this by adopting an objective function that incorporates a policy entropy term as
\begin{equation}
\label{eq_max_ent_policy_objective}
J_{\pi} = \Exp_{\substack{s_t\sim \rho_{\pi}}}\Big[\sum^{\infty}_{t=0}\gamma^{t} [r_t+\alpha\mathcal{H}(\pi(\cdot|s_t))]\Big],
\end{equation}
where $\gamma \in (0,1)$ is the discount factor, $\rho_t$ is the state visitation distribution, $\alpha$ is the temperature coefficient, and the single-step policy entropy $\mathcal{H}$ is expressed as
\begin{equation}
\label{eq_entropy_definition}
\begin{aligned}
\mathcal{H}(\pi(\cdot|s_t))=\Exp_{a_t\sim\pi(\cdot|s_t)}\big[-\log\pi(a_t|s_t)\big].
\end{aligned}
\end{equation}
The optimal policy can be derived through a maximum entropy variant of policy iteration, commonly known as soft policy iteration~\citep{wang2022deep}. This iterative process alternates between two key stages: (1) soft policy evaluation (PEV) and (2) soft policy improvement (PIM). 


In soft PEV, provided a policy $\pi$, for a given policy $\pi$, we can apply the soft Bellman operator $\mathcal{B}^{\pi}$ to learn the soft Q-value, as shown by the soft Bellman expectation equation:
\begin{equation}
\label{eq.soft_bellman}
\begin{aligned}
\mathcal{B}^\text{soft}[Q^{\text{soft}}(s,a)]=r+\gamma \mathbb{E}_{s'\sim p,a'\sim \pi}[Q^{\text{soft}}(s',a')-\alpha \log\pi(a'|s')\big],
\end{aligned}
\end{equation}
where the definition of soft Q-value is
\begin{equation}
Q^{\text{soft}}(s,a) = \mathbb{E}_\pi \left[ \sum_{t=0}^\infty \gamma^t r_t + \sum_{t=1}^\infty \gamma^t \alpha \mathcal{H}(\pi(\cdot |s_t)) \,\bigg|\, s_0 = s, a_0 = a \right].
\end{equation}
One might ask why we write the reward and entropy signals as two separate summation terms. The reason is to highlight the difference in their starting indices. The reward signal is accumulated from the current time step, with a summation index of $t=0$, while the policy entropy is accumulated from the next time step, with a summation index of $t=1$. 
This difference is evident from the soft Bellman expectation equation in \eqref{eq.soft_bellman}: the first term on the right-hand side, $r$, does not have a corresponding policy entropy term at the same time step. In fact, the missing current entropy $\mathcal{H}(\pi(\cdot|s_0))$ occurs in the subsequent soft PIM step.

In soft policy improvement (PIM), the goal is to find a new policy that outperforms the current policy. This is achieved by directly maximizing an entropy-augmented objective, a process equivalent to:
\begin{equation}
\label{eq.policy_imp}
\begin{aligned}
\pi_{\rm{new}}=\arg\max_{\pi} \Exp_{s\sim \rho_{\pi},a\sim \pi}\big[Q^\text{soft}(s,a)-\alpha \log\pi(a|s)\big].
\end{aligned}
\end{equation}
The convergence of soft policy iteration to the optimal maximum entropy policy is a well-established result in the field, as shown by \citep{haarnoja2017reinforcement}.
\paragraph{Temperature tuning.}
A key advancement in the latest maximum entropy frameworks is the automatic management of the {temperature parameter} $\alpha$. Instead of being a fixed hyperparameter, $\alpha$ is treated as a learnable variable.  The objective is to minimize
\begin{equation}
    \label{eq_single_step_entropy_tuning}
   J(\alpha) = \mathbb{E}_{a_t \sim \pi} \big[ -\alpha \big( \log \pi(a_t|s_t) + \mathcal{H}_{0} \big) \big], 
\end{equation}
where the default value of $\mathcal{H}_{0}$ is commonly set as $-\dim(\mathcal{A})$, i.e., the minus of the number of action dimensions. This mechanism achieves a dynamic balance between exploration and exploitation by maintaining the policy's local entropy close to a predefined {target entropy} $\mathcal{H}_{0}$ across all situations~\citep{Haarnoja2018SAC}.

\section{Method}
\label{sec_method}



\subsection{Two Bottlenecks of Maximum Entropy RL}

Previously, we briefly introduced two bottlenecks that exist in the current maximum entropy RL framework. Now, combining with specific formulas, we will more formally and mathematically explain their origins and their impact on policy learning.

\paragraph{(1) Non-stationary Q-value estimation.}
In each soft PEV step, as shown in \eqref{eq.soft_bellman}, the target value is calculated by 
\begin{equation}
\label{eq_issue_non-stationary_q_value}
    y_\text{target} = r(s,a) + \gamma [ Q^{\text{soft}}(s',a') + \alpha \mathcal{H}(\pi(\cdot|s'))].
\end{equation}
When the temperature $\alpha$ is updated at the same time, the target value distribution shifts dynamically. This entanglement injects additional variance and bias into Q-value estimation, degrading subsequent policy improvement steps that rely on stable value predictions.

\paragraph{(2) Short-sighted local entropy tuning.} 
In each soft PIM step, as shown in \eqref{eq_single_step_entropy_tuning}, the existing temperature tuning mechanism aligns every local single-step entropy to a fixed target by adjusting $\alpha$ to match $\mathbb{E}[-\log \pi(a|s)]$ to some desired value.
However, it would be better to adjust the trajectory entropy to control the long-term policy stochasticity, which is defined as:
\begin{equation}
\label{eq_traj_entropy_definition}
    \mathcal{H}_{\text{traj}}(s) = \mathbb{E}_{\tau \sim \pi} \Big[ \sum_{t=0}^\infty \gamma^t \mathcal{H}(\pi(\cdot|s_t)) \bigg|\, s_0 = s \Big].
\end{equation}
In summary, while the maximum-entropy framework is a powerful tool for policy learning, its effectiveness is still hindered by the two identified bottlenecks. These limitations motivate us to execute reward-entropy separation to ensure clean and stable value learning and rethink maximum entropy RL from a trajectory-level entropy constraint perspective.


\subsection{Trajectory Entropy-Constrained Reinforcement Learning}

\begin{figure}
    \centering
    {\includegraphics[ width=0.98\textwidth,trim={1.2cm 0.5cm 0.45cm 0.5cm}, clip]{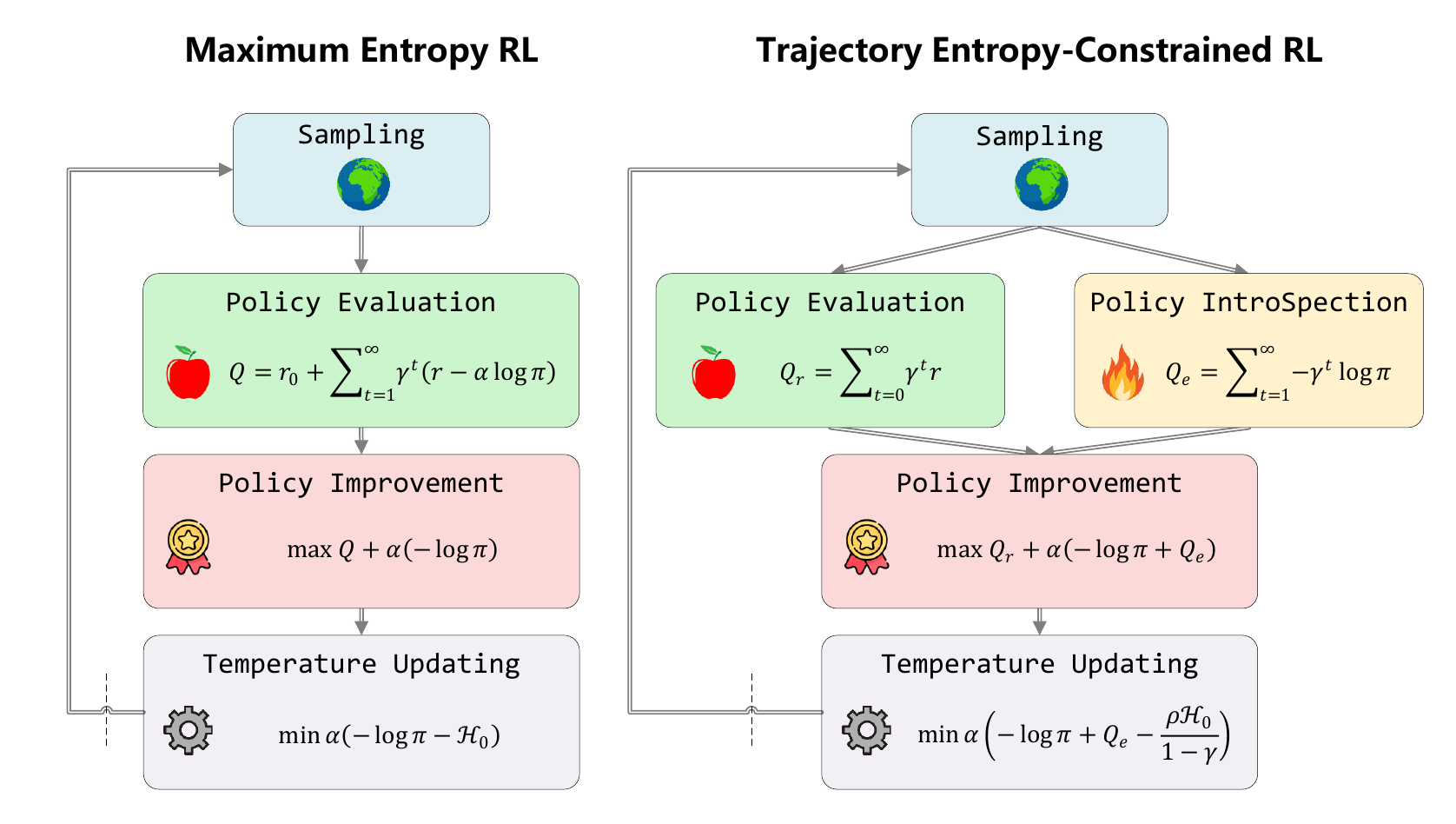}
    }
    \caption{
\textbf{Comparison} between standard maximum entropy RL (\textbf{left}) and our trajectory entropy-constrained (TEC) RL (\textbf{right}). Our TECRL framework comprises four key components: a reward-centric policy evaluation (PEV), an entropy-centric policy introspection (PIS), a policy improvement (PIM) that retains the exact soft policy objective, and a temperature update (TUP) tuning the temperature guided by the trajectory entropy constraint.
} 
    \label{fig:compare}
\end{figure}


To address the two bottlenecks identified earlier, we propose Trajectory Entropy-Constrained Reinforcement Learning (TECRL). It formulates an explicit equality constraint on the trajectory-level entropy to control the policy stochasticity, which yields the following policy optimization problem:
\begin{equation}
\begin{aligned}
 \label{eq_tecrl}
    \max_{\pi} &\; \mathbb{E}_{\pi} \Big[ \sum_{t=0}^\infty \gamma^t [r(s_t,a_t) + \alpha \mathcal{H}(\pi(\cdot|s_t)) \Big] \\
\quad \text{s.t.} &\quad
\mathbb{E}_{\pi} \Big[ \sum_{t=0}^\infty \gamma^t \mathcal{H}(\pi(\cdot|s_t)) \Big] = \mathcal{H}_{\text{budget}}.   
\end{aligned}
\end{equation}
Under this trajectory entropy constraint, the agent is required to strategically distribute a fixed budget of randomness across its entire trajectory. This offers a more principled way to mitigate the dilemma of under- and over-exploration.

 
To practically solve the optimal policy, our TECRL integrates four alternating steps: 
{(1) Policy Evaluation (PEV)} estimates the expected cumulative reward; 
{(2) Policy Introspection (PIS)} estimates the expected cumulative entropy; 
{(3) Policy Improvement (PIM)} jointly leverages both critics to formulate soft policy objective; and 
{(4) Temperature Updating (TUP)} adapts the temperature to enforce the trajectory entropy constraint. Below we detail these four steps one by one.

\paragraph{(1) Policy Evaluation (PEV).}  
This step learns a reward-centric critic $Q_r$ defined as 
\begin{equation}
Q_r(s,a) = \mathbb{E}_\pi \left[ \sum_{t=0}^\infty \gamma^t r_t  \,\bigg|\, s_0 = s, a_0 = a \right],
\end{equation}
The PEV loss follows the standard Bellman expectation equation:
\begin{equation}
\label{eq_pev}
\mathcal{L}_\text{PEV} = (Q_r(s,a)-y_r)^2, \quad \text{where} \quad
  y_r = r(s,a) + \gamma\,\mathbb{E}_{s',a'}[ Q_r(s',a') ], 
\end{equation}
This reward-centric critic explicitly excludes entropy bonuses, which ensures a clean value target uninfluenced by policy stochasticity.

\paragraph{(2) Policy Introspection (PIS).}  
This step learns an entropy-centric critic $Q_e$. For a Gaussian policy, the entropy of the current step is straightforward to compute. Therefore, we define $Q_e$ as the cumulative policy entropy from the next time step to infinity, which is defined as
\begin{equation}
\label{eq_definition_Qe}
Q_{e}(s,a) = \mathbb{E}_\pi \left[\sum_{t=1}^\infty \gamma^t  \mathcal{H}(\pi(\cdot |s_t)) \,\bigg|\, s_0 = s, a_0 = a \right].
\end{equation}
Notably, it also does not contain the temperature $\alpha$, so its target value is clean and explicit. The PIS loss follows an entropy Bellman expectation equation:
\begin{equation}
\label{eq_pis}
\mathcal{L}_\text{PIS} = (Q_e(s,a)-y_e)^2, \quad \text{where} \quad
  y_e = \gamma \mathcal{H}(\pi(\cdot|s')) + \gamma\, Q_e(s',a').
\end{equation}
The mathematical correspondence between \eqref{eq_definition_Qe} and \eqref{eq_pis} can be seen in the Appendix \ref{appendix_Qe_and_bellman}, and the convergence proof of the newly proposed PIS is presented in Appendix \ref{appendix_convergence_pis}.

We refer to this process as policy introspection because the $Q_e$ value reflects the future cumulative entropy of the current policy across different state-action pairs. In essence, it quantifies the long-term stochasticity inherent to the policy itself.

\paragraph{(3) Policy Improvement (PIM).}  
With dual critics $Q_r$ and $Q_e$, We can formulate a policy loss as:
\begin{equation}
\label{eq_pim}
    \mathcal{L}_\text{PIM} =  \underbrace{Q_r(s,a)}_\text{cumulative reward} + ~\alpha \underbrace{(-\log \pi(a|s) + Q_{e}(s,a))}_\text{cumulative entropy} .
\end{equation}
This PIM loss aligns with the soft policy objective shown in \eqref{eq_max_ent_policy_objective}. $Q_r$ represents the cumulative reward, $-\log \pi(a|s)$ is the current policy entropy, and $Q_e$ represents the cumulative entropy starting from the next time step. Therefore, our PIM is compliant with the maximization term in \eqref{eq_tecrl}.

\paragraph{(4) Temperature Updating (TUP).}  
Finally, the aim of TUP is tuning $\alpha$ to enforce the trajectory entropy constraint, whose loss is
\begin{equation}
\label{eq_tup}
\mathcal{L}_\text{TUP} = - \alpha \bigg(\underbrace{-\log \pi(a|s) + Q_{e}(s,a)}_\text{cumulative entropy} - \mathcal{H}_\text{budget}\bigg).
\end{equation}
This mechanism extends existing temperature tuning in \eqref{eq_single_step_entropy_tuning} by replacing uniform local entropy matching with a trajectory-level entropy constraint in \eqref{eq_tecrl}. We set $\mathcal{H}_\text{budget}$ as $\rho H_0 / (1-\gamma)$, The division by $(1-\gamma)$ is to keep the magnitude consistent with the local entropy tuning of the maximum entropy. $\rho$ is an entropy scaling factor that can adjust the budget value.

\paragraph{Summary.}  
Our proposed TECRL framework is grounded in two primary claims:
(1) TECRL enables \textit{more stable and effective exploitation}. This is because the reward-centric value function is now decoupled from the entropy objective, allowing it to provide a more accurate and dedicated prediction to guide policy improvement.
(2) TECRL enables \textit{more strategic and controllable exploration.} By having the agent dynamically allocate its finite entropy budget where it is most needed, the method facilitates the preservation of high-value behaviors while preventing unstable swings in policy stochasticity.
The full pseudocode is summarized in Algorithm~\ref{alg:dsace}.


\begin{algorithm}[t]
\caption{Trajectory Entropy-Constrained Reinforcement Learning (TECRL)}
\label{alg:dsace}
\begin{algorithmic}[1]
\State \textbf{Initialize} policy $\pi_\theta$, reward critic $Q_{r, \psi}$, entropy critic $Q_{e, \phi}$, temperature $\alpha$, replay buffer $\mathcal{D}$
\For{each iteration}
    \State Observe $s_t$, sample $a_t \sim \pi_\theta(a|s_t)$, execute $a_t$, receive $r_t$, next state $s_{t+1}$
    \State Store $(s_t,a_t,r_t,s_{t+1})$ in $\mathcal{D}$
    \State Sample mini-batch $\{(s,a,r,s')\} \sim \mathcal{D}$
    \State Update $Q_r$ with \eqref{eq_pev} \Comment{(PEV) Policy Evaluation}
    \State Update $Q_e$ with  \eqref{eq_pis} 
     \Comment{(PIS) Policy Introspection}
    \State Update $\pi_\theta$ with   \eqref{eq_pim} \Comment{(PIM) Policy Improvement}
    \State Update $\alpha$ with  \eqref{eq_tup} 
     \Comment{(TUP) Temperature Updating}
\EndFor
\end{algorithmic}
\end{algorithm}

\subsection{Theoretical Analysis on Performance Bound}

We formalize how a trajectory entropy constraint affects policy performance and demonstrate why a properly chosen entropy budget can raise the performance upper bound.
We first denote $\pi_{\text{soft}}^*$ as the optimal policy under the standard maximum entropy RL setting, which maximizes the soft objective
\begin{equation}
J_{\text{MaxEnt}}(\pi_{\text{soft}}^*) = R_{\text{MaxEnt}}^* + \alpha_{\text{soft}}^*\,\mathcal{H}_{\text{soft}}^*,
\end{equation}
where $R_{\text{MaxEnt}}^*$ and $\mathcal{H}_{\text{soft}}^*$ represent the optimal return and cumulative entropy, respectively, and $\alpha_{\text{soft}}^*>0$ is the optimal temperature parameter.

Let $R_\text{TEC}$ be the return of our TECRL policy. We assume that the entropy budget $\mathcal{H}_{\text{budget}}$ is chosen to be within the feasible range of entropy values encountered during the MaxEnt optimization process. Specifically, it is neither smaller than the minimal achievable entropy nor larger than the maximal entropy $\mathcal{H}_{\text{soft}}^*$ obtained by the optimal maximum-entropy policy $\pi_{\text{soft}}^*$. Therefore, with the same temperature $\alpha_{\text{soft}}^*$, we have the following inequality
\begin{equation}
\label{theo_inequ}
J_{\text{MaxEnt}}(\pi_{\text{soft}}^*) \ge R_\text{TEC} + \alpha_{\text{soft}}^*\,\mathcal{H}_{\text{budget}}^*.
\end{equation}
By rearranging this inequality, the return of our TECRL can be bounded from above as
\begin{equation}
\begin{aligned}
R_\text{TEC} & \leq J_{\text{MaxEnt}}(\pi_{\text{soft}}^*) - \alpha_{\text{soft}}^*\,\mathcal{H}_{\text{budget}} \\
& = R_{\text{MaxEnt}}^* + \alpha_{\text{soft}}^*\,(\mathcal{H}_{\text{soft}}^* - \mathcal{H}_{\text{budget}}).
\end{aligned}
\end{equation}
This inequality explicitly shows that our achievable return is bounded by a quantity proportional to the entropy gap $\mathcal{H}_{\text{soft}}^* - \mathcal{H}_{\text{budget}}$. This analysis demonstrates that appropriately selecting a trajectory entropy budget can lead to a higher performance bound.

\section{Experiments}
\label{sec_experiment}

\subsection{Main Experiment}
\paragraph{Benchmark.}  We evaluate performance on a suite of standard continuous control tasks from the OpenAI Gym interface~\citep{brockman2016openaigym}. Specifically, we choose 8 Mujoco tasks: Humanoid-v3, Ant-v3, Hopper-v3, Walker2d-v3, Swimmer-v3, HalfCheetah-v3, InvertedDoublePendulum-v2 (abbreviated as InvertedDP-v2) and Reacher-v2. Details are provided in Appendix \ref{appendix_env}.


\paragraph{Baselines.} We consider 7 well-known model-free algorithms, including trust region policy optimization (TRPO) \citep{schulman2015TRPO}, proximal policy optimization (PPO) \citep{schulman2017PPO}, deep deterministic policy gradient (DDPG) \citep{lillDDPGicrap2015DDPG}, twin delayed deep deterministic policy gradient (TD3) \citep{Fujimoto2018TD3}, soft actor-critic (SAC) \citep{Haarnoja2018SAC}, Distributional SAC (DSAC)~\citep{duan2021distributional} and its latest version DSAC-T~\citep{duan2025distributional}.  See Appendix \ref{appen_hyper} for detailed hyperparameters. 



\begin{figure*}[t]
\centering
\captionsetup[subfigure]{justification=centering}
\subfloat[Humanoid-v3\label{subFig:humanoid}]
{\includegraphics[width = 0.25\textwidth]{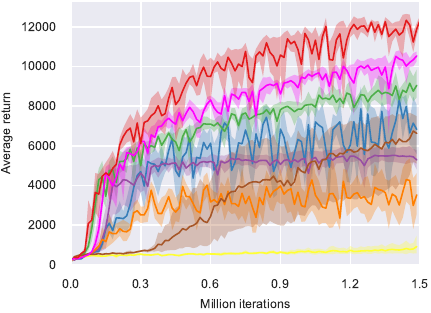}} 
\subfloat[Ant-v3\label{subFig:ant}]
{\includegraphics[width = 0.25\textwidth]{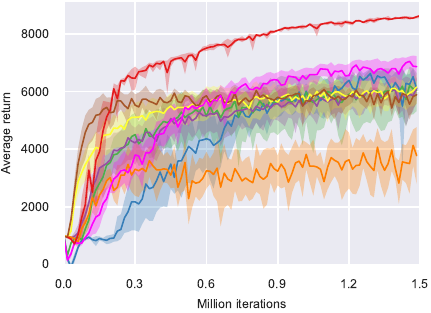}} 
\subfloat[Hopper-v3\label{subFig:hopper}]
{\includegraphics[width = 0.25\textwidth]{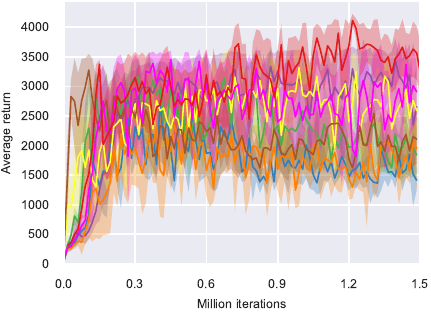}} 
\subfloat[Walker2d-v3\label{subFig:walker2d}]
{\includegraphics[width = 0.25\textwidth]{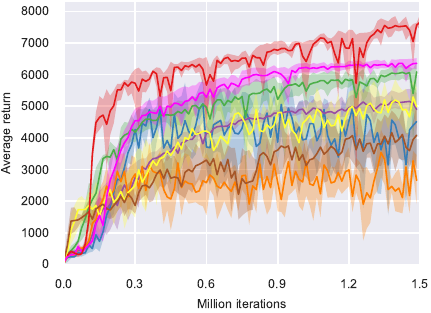}} \\
\subfloat[Swimmer-v3\label{subFig:idp}]
{\includegraphics[width = 0.25\textwidth]{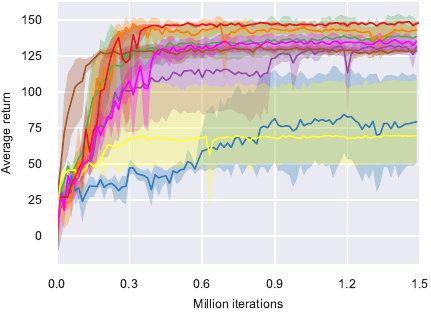}} 
\subfloat[Halfcheetah-v3\label{subFig:halfcheetah}]
{\includegraphics[width = 0.25\textwidth]{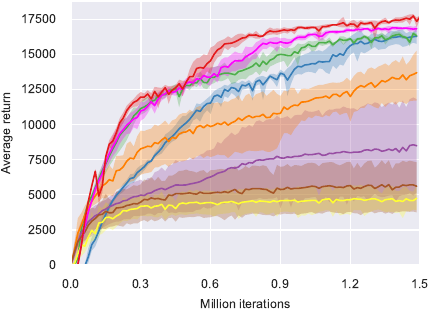}} 
\subfloat[InvertedDP-v2\label{subFig:IDP}]
{\includegraphics[width = 0.25\textwidth]{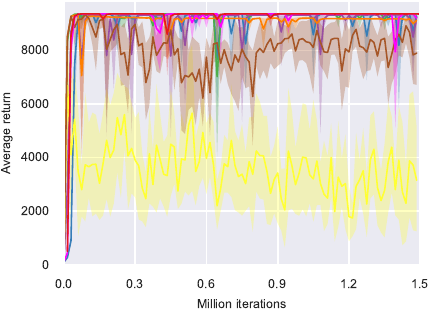}} 
\subfloat[Reacher-v2\label{subFig:reacher}]
{\includegraphics[width = 0.25\textwidth]{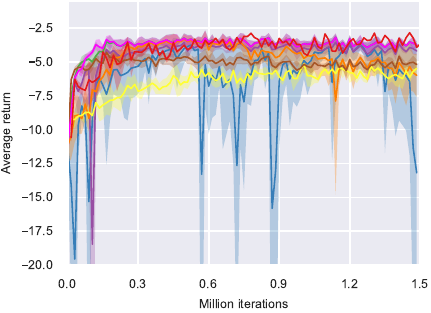}}
\\
 \subfloat
{\includegraphics[width = 0.98\textwidth,trim={0.5cm 0.0cm 0.5cm 0.0cm}, clip]{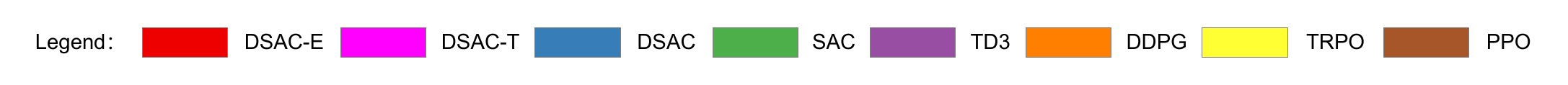}} 
\caption{\textbf{Training curves on benchmarks.} The solid lines correspond to mean and shaded regions
correspond to the 95\% confidence interval over five runs.}
\label{fig_benchmeark}
\end{figure*}

\begin{table*}[!htb]
\setlength{\tabcolsep}{4pt} 
 \centering
    \caption{\textbf{Average final return.} Computed as the mean of the highest return values observed in the final 10\% of iteration steps per run. $\pm$ corresponds to standard deviation over 5 runs.}
\label{tab_benchmark}
\begin{tabular}{l|l|l| c c c c}
  \toprule
  \multicolumn{3}{c|}{Algorithm} & Humanoid-v3 & Ant-v3 &  Hopper-v3& Walker2d-v3\\ 
  \midrule
  \multirow{6}{*}{\shortstack{Off}} & \multirow{4}{*}{\shortstack{w/ entropy}}& DSAC-E & \cellcolor{red!25} {\textbf{12542$\pm$280}} 
  & \cellcolor{red!25}\textbf{8640$\pm$57}~~ 
  & \cellcolor{red!25}\textbf{3901$\pm$385} 
  & \cellcolor{red!25}\textbf{7780$\pm$137} 
  \\
  &&DSAC-T & \cellcolor{ProcessBlue!25}{10829$\pm$243} & \cellcolor{ProcessBlue!25}{7086$\pm$261} & \cellcolor{ProcessBlue!25}{3660$\pm$533} & \cellcolor{ProcessBlue!25}{6424$\pm$147}\\
  &&DSAC & ~~9074$\pm$286 & 6862$\pm$53~~ & 2135$\pm$434 & 5413$\pm$865\\
  &&SAC & ~~9336$\pm$696 & 6427$\pm$805 & 2483$\pm$943 & 6201$\pm$264\\
  \cmidrule(r){2-7}
  &\multirow{2}{*}{\shortstack{w/o entropy}}  &TD3 & ~~5632$\pm$436 & 6184$\pm$487 & 3569$\pm$455 & 5238$\pm$336\\
  &&DDPG & ~~5292$\pm$663 & 4549$\pm$789 & 2644$\pm$659 & 4096$\pm$68~~\\
  \cmidrule(r){1-7}
  \multirow{2}{*}{On}& \multirow{2}{*}{w/ entropy} &TRPO & ~~~~965$\pm$555 & 6203$\pm$579 & 3474$\pm$400 & 5503$\pm$593\\ 
  &&PPO & ~~~~6869$\pm$1563 & 6157$\pm$185 & 2647$\pm$482 & 4832$\pm$638\\
  \midrule
    &&$\Uparrow$ & \cellcolor{Green!15}15.82\% & \cellcolor{Green!15}21.93\% & \cellcolor{Green!15}6.58\% & \cellcolor{Green!15}21.11\%\\
  \midrule
  \midrule
  \multicolumn{3}{c|}{Algorithm}  & Swimmer-v3& Halfcheetah-v3 & InvertedDP-v2 & Reacher-v2 \\
  \midrule
  \multirow{6}{*}{\shortstack{Off}} & \multirow{4}{*}{\shortstack{w/ entropy}}& DSAC-E & \cellcolor{red!25}$\textbf{149.3$\pm$0.3}$~~ &  \cellcolor{red!25}$\textbf{17904$\pm$100}$ &  \cellcolor{red!25}$\textbf{9360$\pm$0}$ & \cellcolor{red!25}$\textbf{-2.9$\pm$0.1}$ \\
  &&DSAC-T  & 137.6$\pm$6.4~~~& \cellcolor{ProcessBlue!25}17025$\pm$157 &  \cellcolor{red!25}$\textbf{9360$\pm$0}$ & \cellcolor{ProcessBlue!25}-3.1$\pm$0.2\\
  &&DSAC  & ~~83.9$\pm$35.6 & 16542$\pm$514 &  \cellcolor{ProcessBlue!25}{9359$\pm$1} & -4.3$\pm$1.9\\
  &&SAC  & 140.4$\pm$14.3 & 16573$\pm$224 &  \cellcolor{red!25}$\textbf{9360$\pm$0}$ & \cellcolor{ProcessBlue!25}-3.1$\pm$0.2 \\
  \cmidrule(r){2-7}
  &\multirow{2}{*}{\shortstack{w/o entropy}}  &TD3  & 134.0$\pm$5.4~~ & ~~~~8633$\pm$4041 & ~~9347$\pm$15 & -3.4$\pm$0.2 \\
  &&DDPG  & \cellcolor{ProcessBlue!25}145.6$\pm$4.3~~ &~~13970$\pm$2083 & ~~9183$\pm$10 & -4.5$\pm$1.3 \\
  \cmidrule(r){1-7}
  \multirow{2}{*}{On}& \multirow{2}{*}{w/ entropy} &TRPO  & ~~70.4$\pm$38.1& ~~4785$\pm$968 & ~~~~~~6260$\pm$2066 & -5.0$\pm$0.6\\ 
  &&PPO  & ~130.3$\pm$2.0~~~~& ~~~~5790$\pm$2201 & 9357$\pm$2 & -4.4$\pm$0.2\\
    \midrule
    &&$\Uparrow$ & \cellcolor{Green!15}2.54\% & \cellcolor{Green!15}5.16\% & \cellcolor{Green!15}0\% & \cellcolor{Green!15}6.45\%\\
  \bottomrule
\end{tabular}
\\[2mm]
\begin{minipage}{\textwidth}
\footnotesize
~~~~~~ * {\textbf{Bolded}} and red = best, blue = second-best. \textit{$\Uparrow$ means the improvement of the best over the second-best.}
\end{minipage}
\end{table*}




\paragraph{Our method.} Our proposed DSAC-E algorithm is built on the DSAC-T, inheriting all of its hyperparameters. For the newly introduced hyperparameter $\rho$, we set its value to 20 for the Humanoid-v3 and Walker2d-v3 tasks, and to 1 for all other tasks. 
The reason for setting larger $\rho$ values for these two tasks is that they are relatively high-dimensional and that the robots are particularly prone to falling over due to overly random actions.  Recall that the base single-step entropy budget $H_0$ is a negative value, so a larger $\rho$ means a smaller budget $\rho H_0 / (1-\gamma)$.

\paragraph{Evaluation protocol.}
 The total training step  for all experiments is set at 1.5 million, with the results of all
experiments averaged over 5 random seeds.
For each seed,
the metric is derived by averaging the highest return values observed during the final 10\% of iteration
steps in each run, with evaluations conducted every 15,000 iterations. 
Each assessment result is the
average of ten episodes. The results from the 5 seeds are then aggregated to calculate the mean and
standard deviation.



\paragraph{Main results.} 
 Figure \ref{fig_benchmeark} and Table \ref{tab_benchmark} display all the learning curves and numerical performance results
, respectively. Our comprehensive findings reveal that across all evaluated 8 tasks, the DSAC-E algorithm consistently matched or surpassed the performance of all competing benchmark algorithms, establishing new state-of-the-art results. Notably, it achieved less oscillation and substantial performance improvements on the Humanoid-v3, Ant-v3, Walker2d-v3, and Hopper-v3 tasks, with improvements of 15.82\%, 21.93\%, 21.11\% and 6.6\% over the second-best. 

\subsection{Ablation Study}
We conduct ablation studies on the Humanoid-v3 task to evaluate the contribution of each component.

\begin{figure}[htbp!]
\centering
\begin{minipage}[t]{0.45\textwidth}
    \centering
    \includegraphics[width = 0.85\textwidth]{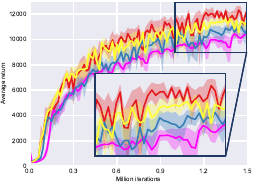}
    \caption{Ablation on the TEC and RES.}
    \label{fig_ablation1}
\end{minipage}%
\begin{minipage}[t]{0.5\textwidth}
    \centering
    \vspace{-110pt}
    \includegraphics[width = 0.9\textwidth,trim={0cm 0.3cm 0.5cm 0.3cm}, clip]{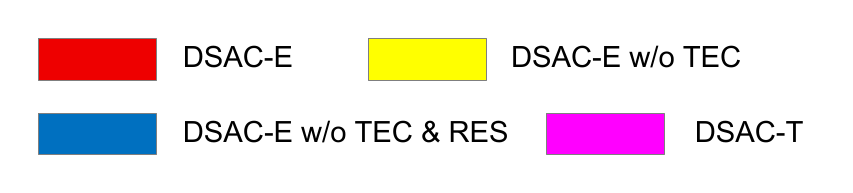}
    \captionof{table}{Results of ablation on TEC and RES.}
    \label{tab_ablation1}
    \begin{tabular}{l c}
    \toprule
    Algorithm & TAR \\
    \midrule
    DSAC-E (full)  &  $\textbf{12542 $\pm$ 280}$  \\
    DSAC-E w/o TEC &  {11786 $\pm$ 374}  \\
    DSAC-E w/o TEC \& RES & 11455 $\pm$ 404 \\
    DSAC-T & 10829 $\pm$ 243 \\
    \bottomrule
    \end{tabular}
\end{minipage}
\end{figure}

\paragraph{Reward-entropy separation (RES) and trajectory Entropy Constraint (TEC).}
We perform a step-wise ablation, considering four algorithms:
(1) Our full DSAC-E.
(2) DSAC-E w/o TEC, which replaces our trajectory entropy constraint with existing local entropy tuning.
(3) DSAC-E w/o TEC and RES, which is close to DSAC-T but with a $\rho$ value of 20.
(4) original DSAC-T, which can be understood as having a $\rho$ of 1.
As shown in Figure \ref{fig_ablation1} and Table \ref{tab_ablation1}, the performance of the algorithms progressively declines as more components are removed. This result confirms the effectiveness of both our RES and TEC modules. Next we will provide a more systematic analysis of the $\rho$.

\begin{figure}[h]
\centering
\captionsetup[subfigure]{justification=centering}
\subfloat[DSAC-T\label{subFig:ablation_T}]
{\includegraphics[height = 0.3\textwidth,trim={0.0cm 0.0cm 0.0cm 0.0cm}, clip]{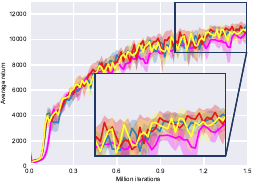}} 
\subfloat[DSAC-E (ours)\label{subFig:ablation_E}]
{\includegraphics[height = 0.3\textwidth,trim={0.0cm 0.0cm 0.0cm 0.0cm}, clip]{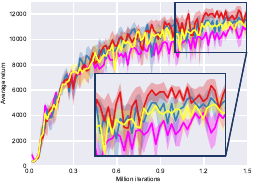}}\\
 \includegraphics[width = 0.72\textwidth,trim={0.5cm 0.5cm 0.0cm 0.0cm}, clip]{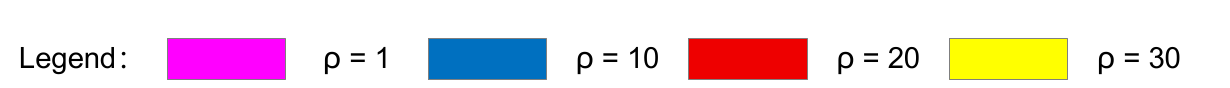}
\caption{Ablation on the sensitivity to the trajectory entropy budget. 
    }
\label{fig_ablation2}
\end{figure}

\paragraph{Impact of $\rho$ controlling trajectory entropy budget.}
We further investigate the effect of varying the trajectory entropy budget. Specifically, we apply different $\rho$ values for both DSAC-T and our DSAC-E.
As shown in Figure \ref{fig_ablation2} and Table \ref{tab_ablation2}, the performance gain of DSAC-T (Figure \ref{subFig:ablation_T})  is not significant with the adjustment of  $\rho$.
Its performance varies only slightly and all results cluster closely together. In contrast, our DSAC-E (Figure \ref{subFig:ablation_E}) consistently outperforms DSAC-T across all settings, and its performance shows a clearer, more structured dependence on $\rho$.



For both DSAC-T and our DSAC-E, performance first improves and then degrades as $\rho$ increases, which aligns with our theoretical analysis: a properly chosen entropy budget can lift the performance bound, whereas an excessively large $\rho$ (corresponding to an overly small entropy budget) reduces exploration and leads to a performance drop. Overall, our DSAC-E achieves higher performance and exhibits a more interpretable sensitivity to $\rho$, making it easier to tune for high returns.
\begin{table}[h]
\centering
\caption{Performance of DSAC-T and our DSAC-E under different $\rho$ values.}
\label{tab_ablation2}
\begin{tabular}{l c c c c}
\toprule
Algorithm & $\rho=1$ & $\rho=10$ & $\rho=20$ & $\rho=30$ \\
\midrule
DSAC-T & 10829 $\pm$ 243 & 11079 $\pm$ 457 & 11455 $\pm$ 404 & 11182 $\pm$ 705 \\
DSAC-E (ours) & \textbf{11382 $\pm$ 447} & \textbf{12118 $\pm$ 505} & \textbf{12542 $\pm$ 280} & \textbf{11747 $\pm$ 365} \\
\bottomrule
\end{tabular}
\end{table}
\section{Related Work}
\label{sec_related_work}
Exploration remains a central challenge in RL, and prior studies have proposed various strategies to inject and regulate stochasticity into the policy~\citep{amin2021survey}. Broadly, existing approaches can be grouped into two main categories: action-noise-based and maximum-entropy-based exploration~\citep{hao2023exploration}. While other alternatives, such as curiosity-driven~\citep{sun2022exploit} or uncertainty-based~\citep{an2021uncertainty} exploration, have been explored, they remain less commonly adopted in standard model-free RL algorithms.
\paragraph{Action-noise based exploration.}  
A line of methods in off-policy RL encourages exploration by directly perturbing the agent’s actions with a noise process. For instance, DDPG first \citep{lillDDPGicrap2015DDPG} employs Ornstein–Uhlenbeck noise to facilitate temporally correlated exploration, and the TD3 family \citep{Fujimoto2018TD3, fujimoto2023sale, seo2025fasttd3} turn to simply apply Gaussian noise to each action dimension to effectively maintain randomness during training. Although these approaches are intuitive and easy to implement, they suffer from two key drawbacks. First, the noise is added externally and is entirely separate from the policy's learning objective. The policy itself is unaware of this exploration mechanism, making it a blind, ad hoc process~\citep{plappert2018parameter, li2021adaptive}. Second, it creates a fundamental inconsistency between training and evaluation. A policy trained with exploratory noise is different from the final policy used for deployment, which can lead to a policy-value mismatch and hinder convergence to a truly optimal solution~\citep{hollenstein2022action, sikchi2022learning}. Overall, although action-noise based exploration is straightforward to implement and can yield good performance, its largely heuristic nature diminishes its reliability.

\paragraph{Maximum-entropy based exploration.}  
A more principled framework for exploration is provided by maximum-entropy RL \citep{haarnoja2017reinforcement}. By augmenting the standard RL objective with an entropy term, methods such as SAC \citep{Haarnoja2018SAC} optimize for both expected return and policy entropy, thereby encouraging diverse behaviors~\citep{nachum2017bridging}. While the latest extensions of SAC further incorporate distributional critics to improve performance~ \citep{duan2021distributional, duan2025distributional}, they share the same tuning principle of maintaining the policy's single-step entropy at a fixed target. Recent work has explored the use of generative models, such as diffusion models, as policy functions~\citep{yang2023diffusion, zhu2023diffusion}. While it's difficult to accurately compute the entropy of this class of functions~\citep{yang2023policy}, these methods still try to follow the standard maximum-entropy principle and entropy tuning mechanism for exploration, for example, by approximating the policy entropy via GMM fitting or alternatively optimizing the lower bound~\citep{wang2024diffusion, wang2025enhanced, ding2024diffusion, celik2025dime}.
Their entropy tuning mechanism remains inherently uniform across all situations and does not explicitly account for long-term policy stochasticity and the inherent need for adaptive exploration.
Our TECRL also employs entropy to monitor policy's stochasticity. However, we shift the focus from local entropy tuning to \emph{trajectory entropy constraint}, highlighting a new perspective on managing policy's long-term stochasticity.
We believe this work provides a new avenue for better resolving the exploitation-exploration dilemma, leading to higher performance.



\section{Conclusion}
\label{sec_conclusion}
In this paper, we revisit the standard maximum entropy RL framework and introduce the trajectory entropy-constrained reinforcement learning (TECRL) framework. Our work addresses two key limitations: (1) non-stationary Q-value estimation and (2) short-sighted local entropy tuning. By separating the reward and entropy Q-functions and applying the trajectory entropy constraint, our framework ensures stable value targets and effective control of long-term policy stochasticity.
Building on this, we develop a practical algorithm, DSAC-E, which extends the state-of-the-art DSAC-T baseline. Empirical results on the OpenAI Gym benchmark show that DSAC-E achieves superior returns and greater stability, validating the effectiveness of our TECRL framework.   

Moving forward, we plan to validate the applicability of our TECRL framework to real-world robotics and large language models (LLMs). This integration will allow agents to benefit from TECRL's superior long-term stochasticity management, leading to more effective and robust behaviors. We believe this work offers a promising paradigm for addressing the exploration-exploitation trade-off and paving the way for more powerful and robust RL agents.

\bibliography{iclr2026_conference}
\bibliographystyle{iclr2026_conference}

\newpage
\appendix
\section{Theoretical Analysis}

\subsection{Useful Lemmas}
\begin{lemma}[Convergence of \(\gamma\)-Contraction Mappings]\label{gamma_contraction_mapping}
Let \((X, d)\) be a complete metric space, and let \(\mathcal{B}: X \to X\) be a \(\gamma\)-contraction mapping with \(0 < \gamma < 1\). This means that for all \(x, y \in X\),
\begin{equation}
d(\mathcal{B}(x), \mathcal{B}(y)) \leq \gamma \cdot d(x, y),
\end{equation}
where \(d\) is the metric on \(X\). According to Banach’s fixed-point theorem, \(\mathcal{B}\) has a unique fixed point \(x^* \in X\), such that \(\mathcal{B}(x^*) = x^*\). Furthermore, for any initial point \(x_0 \in X\), the iterative sequence \(\{x_n\}\) defined by \(x_{n+1} = \mathcal{B}(x_n)\) converges to \(x^*\). The convergence rate is geometric, and we have the inequality
\begin{equation}
d(x_n, x^*) \leq \gamma^n \cdot d(x_0, x^*), \quad \forall n \geq 0.
\end{equation}
This result not only guarantees the existence and uniqueness of the fixed point but also provides a precise rate at which the sequence approaches \(x^*\), demonstrating the efficiency of contraction mappings in finding fixed points.
\end{lemma}

\subsection{Entropy Bellman Expectation equation in Policy Introspection (PIS)}
\label{appendix_Qe_and_bellman}
Here, we build the correspondence between the definition of $Q_e$ in \eqref{eq_definition_Qe} and the entropy Bellman expectation equation in \eqref{eq_pis}.

$Q_e$ represents the cumulative policy entropy starting from the next time step, expressed as:

\begin{equation}
Q_{e}(s,a) = \mathbb{E}_\pi \left[\sum_{t=1}^\infty \gamma^t  \mathcal{H}(\pi(\cdot |s_t)) \,\bigg|\, s_0 = s, a_0 = a \right].
\end{equation}

Our proposed entropy Bellman expectation equation in \eqref{eq_pis} states
\begin{equation}
Q_e(s,a) = \gamma \mathcal{H}(\pi(\cdot|s')) + \gamma\, Q_e(s',a').
\label{entropy_bellman}
\end{equation}




Substitute the definition of $Q_e$ into the RHS of \eqref{eq_pis}, we have:

\begin{equation}
\begin{aligned}
    RHS &= \gamma \mathcal{H}(\pi(\cdot|s')) + \gamma\, Q_e(s',a') \\
    & = \gamma \mathcal{H}(\pi(\cdot|s_{1})) + \gamma \sum_{t=1}^\infty \gamma^t  \mathcal{H}(\pi(\cdot |s_{t+1})) \\
    & = \gamma \mathcal{H}(\pi(\cdot|s_{1})) +  \sum_{t=1}^\infty \gamma^{t+1} \mathcal{H}(\pi(\cdot |s_{t+1})) \\
    & = \gamma \mathcal{H}(\pi(\cdot|s_{1})) +  \sum_{t=2}^\infty \gamma^{t} \mathcal{H}(\pi(\cdot |s_{t})) \\
    & =  \sum_{t=1}^\infty \gamma^{t} \mathcal{H}(\pi(\cdot |s_{t})) = LHS.\\
\end{aligned}
\end{equation}

Thus, we have proven that the definition of $Q_e$ is the solution of the entropy Bellman expectation equation.

\subsection{Convergence of Policy Introspection (PIS)}
\label{appendix_convergence_pis}

We prove the convergence of PIS by showing that the entropy Bellman operator \(\mathcal{B}_{e}\), defined as
\begin{equation}
\label{eq.soft_bellman_e}
\begin{aligned}
\mathcal{B}_{e}Q_e(s,a)=\gamma [Q_e(s',a')-\alpha \log\pi(a'|s')\big],
\end{aligned}
\end{equation}
is a \(\gamma\)-contraction mapping.

We analyze the infinity norm of \(\mathcal{B}_{e}\). For any two functions \(Q_{e, 1}(s, a)\) and \(Q_{e, 2}(s, a)\), we have:
\begin{equation}
\begin{aligned}
\Vert \mathcal{B}_{e}[Q_{e, 1}(s, a)] - \mathcal{B}_{e}[Q_{e, 2}(s, a)] \Vert_\infty 
&= \Vert \gamma [Q_{e, 1}(s',a')-\alpha \log\pi(a'|s')\big] \\
& \hspace{20pt} - \gamma [Q_{e, 2}(s',a')-\alpha \log\pi(a'|s')\big] \Vert_\infty \\
&\leq \Vert \gamma Q_{e, 1}(s',a') - \gamma Q_{e, 2}(s',a') \Vert_\infty \\
&= \gamma \Vert Q_{e, 1}(s',a') - Q_{e, 2}(s',a') \Vert_\infty.
\end{aligned}
\end{equation}

Since \(\gamma \in (0, 1)\), it follows that \(\mathcal{B}_{e}\) is a \(\gamma\)-contraction mapping. By applying Lemma \ref{gamma_contraction_mapping}, we know that \(\mathcal{B}_{e}\) has a unique fixed point. This fixed point can be obtained by iteratively applying \(\mathcal{B}_{e}\) starting from an arbitrary initial  \(Q_{e, \text{init}}\). That is, as the iteration number \(k\) increases, the sequence of updated Q functions converges to a fixed point, i.e., the desired $Q_e$.

\section{Environmental Introduction}
\label{appendix_env}

\paragraph{MuJoCo:} This is a high-performance physics simulation platform widely adopted for robotic reinforcement learning research. The environment features efficient physics computation, accurate dynamic system modeling, and comprehensive support for articulated robots, making it an ideal benchmark for RL algorithm development. 

\begin{figure}[h]
\centering
\captionsetup[subfigure]{justification=centering}
\subfloat[\label{subFig:envhumanoid}]
{\includegraphics[width = 0.24\textwidth, height = 0.24\textwidth]{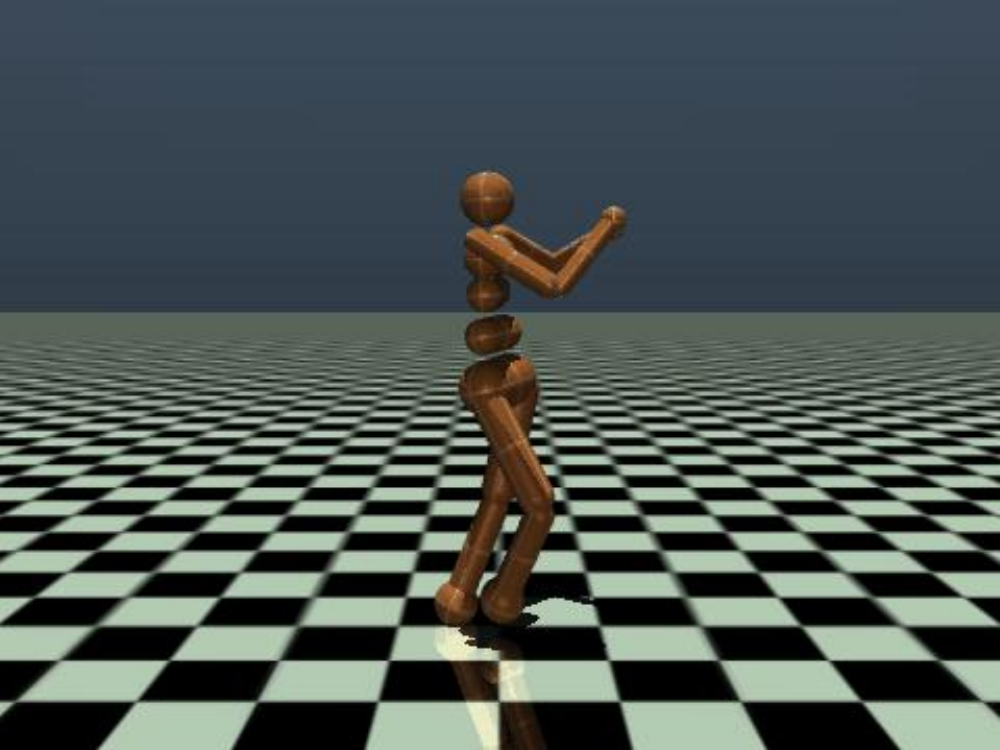}} 
\subfloat[\label{subFig:envant}]
{\includegraphics[width = 0.24\textwidth, height = 0.24\textwidth]{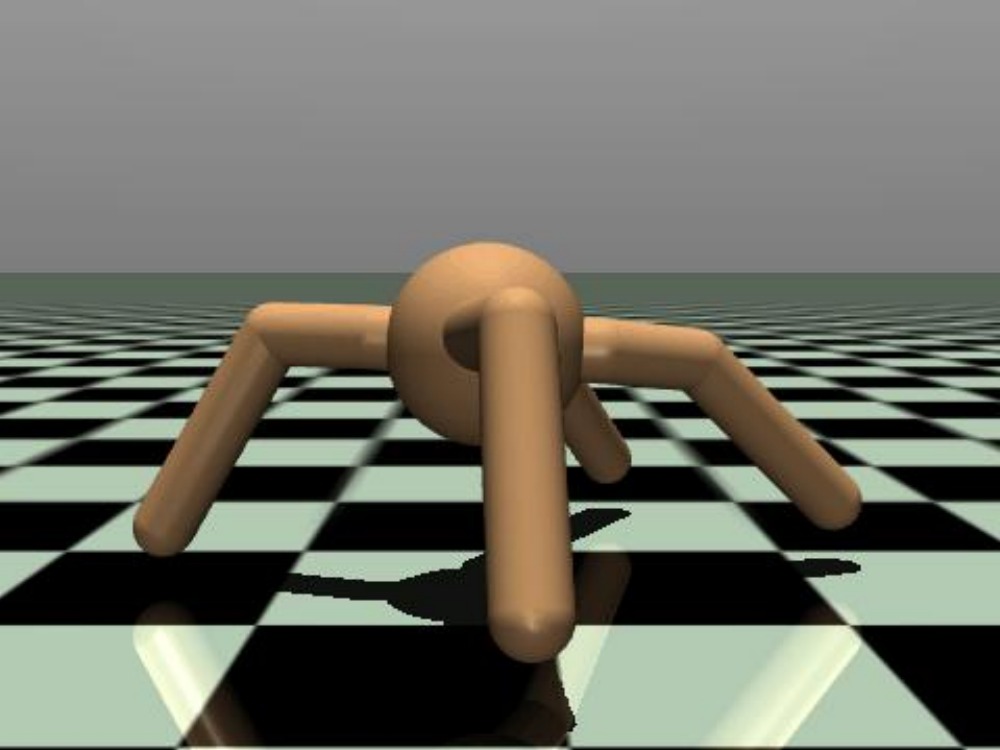}}
\subfloat[\label{subFig:envhalfcheetah}]
{\includegraphics[width = 0.24\textwidth, height = 0.24\textwidth]{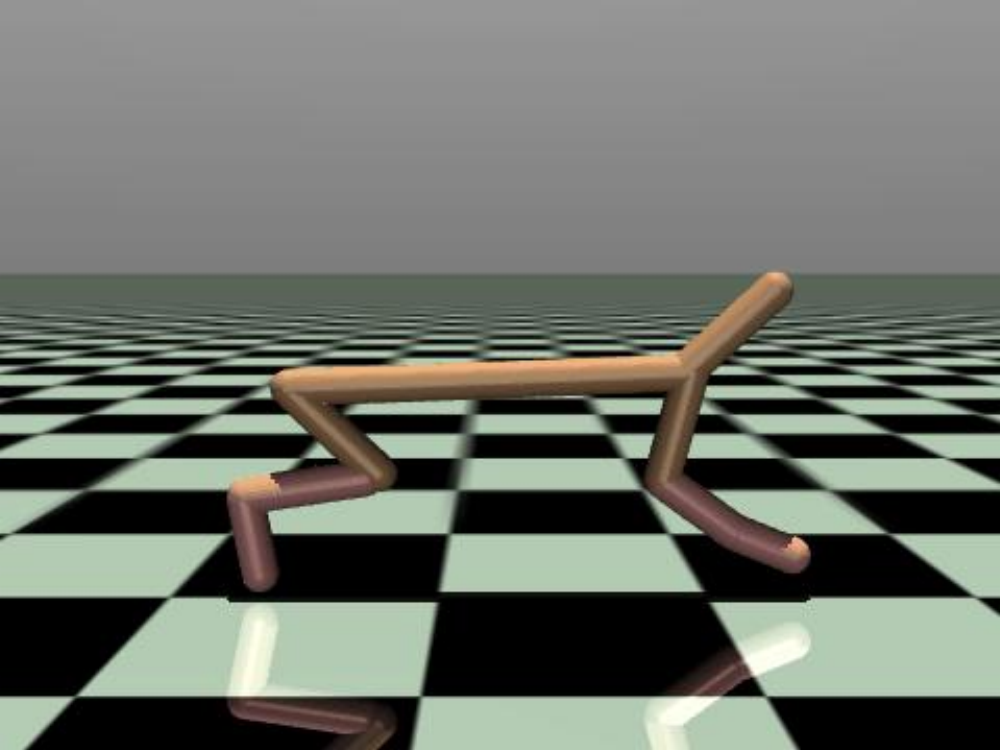}}
\subfloat[\label{subFig:envwalker2d}]
{\includegraphics[width = 0.24\textwidth, height = 0.24\textwidth]{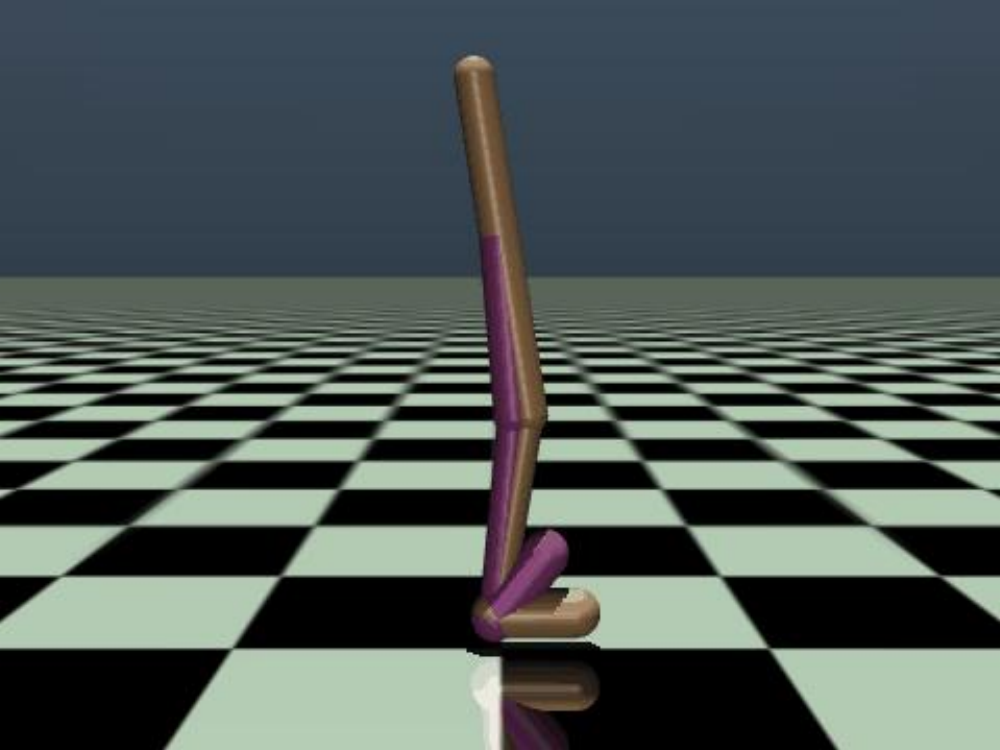}} \\
\subfloat[\label{subFig:envhopper}]
{\includegraphics[width = 0.24\textwidth, height = 0.24\textwidth]{figure/hopper.pdf}}
\subfloat[\label{subFig:envidp}]
{\includegraphics[width = 0.24\textwidth, height = 0.24\textwidth]{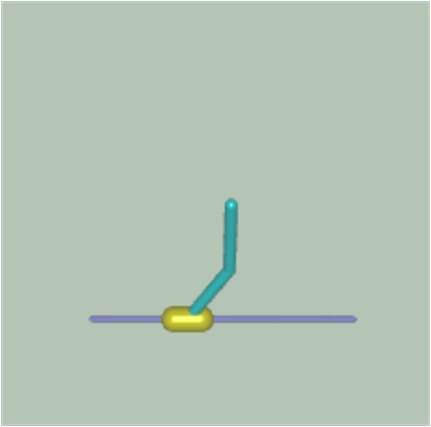}}
\subfloat[\label{subFig:envreacher}]
{\includegraphics[width = 0.24\textwidth, height = 0.24\textwidth]{figure/reacher.pdf}} 
\subfloat[\label{subFig:envswimmer}]
{\includegraphics[width = 0.24\textwidth, height = 0.24\textwidth]{figure/swimmer.pdf}} 

\caption{Benchmarks. 
    (a) Humanoid-v3: \((s \times a) \in \mathbb{R}^{376} \times \mathbb{R}^{17}\). 
    (b) Ant-v3: \((s \times a) \in \mathbb{R}^{111} \times \mathbb{R}^{8}\).
    (c) HalfCheetah-v3: \((s \times a) \in \mathbb{R}^{17} \times \mathbb{R}^{6}\).
    (d) Walker2d-v3: \((s \times a) \in \mathbb{R}^{17} \times \mathbb{R}^{6}\).
    (e) Hopper-v3: \((s \times a) \in \mathbb{R}^{11} \times \mathbb{R}^{3}\).
    (f) InvertedDoublePendulum-v2: \((s \times a) \in \mathbb{R}^{6} \times \mathbb{R}^{1}\).
    (g) Reacher-v2: \((s \times a) \in \mathbb{R}^{11} \times \mathbb{R}^{2}\).
    (h) Swimmer-v3: \((s \times a) \in \mathbb{R}^{8} \times \mathbb{R}^{2}\).
    }

\label{fig_envs}
\end{figure}

In this paper, we concentrate on eight tasks: Humanoid-v3, Ant-v3, HalfCheetah-v3, Walker2d-v3, InvertedDoublePendulum-v3 (InvertedDP-v2), Hopper-v3, Reacher-v2, and Swimmer-v3, as illustrated in Figure \ref{fig_envs}. The InvertedDP-v3 task entails maintaining the balance of a double pendulum in an inverted state. In contrast, the objective of the other tasks is to maximize the forward velocity while avoiding falling. All these tasks are realized through the OpenAI Gym interface \citep{brockman2016openaigym}. 

\clearpage
\section{Visualizations}
To demonstrate the effectiveness of DSAC-E in solving complex, high-dimensional locomotion tasks, we provide visualizations of policy control process on three of the most challenging benchmarks in the Humanoid task as shown in the following Figure \ref{fig_video_frames}. These tasks require precise coordination across many degrees of freedom and long-horizon reasoning. 

The visualization showcase that DSAC-E not only achieves successfully running but also learns robust posture and behaviors, highlighting its strong  capabilities in difficult control scenarios.

\begin{figure}[h]
\centering
\captionsetup[subfigure]{justification=centering}
\subfloat[\textbf{DSAC-E} step 70\label{subFig:dsac1}]
{\includegraphics[width = 0.2\textwidth, height = 0.24\textwidth]{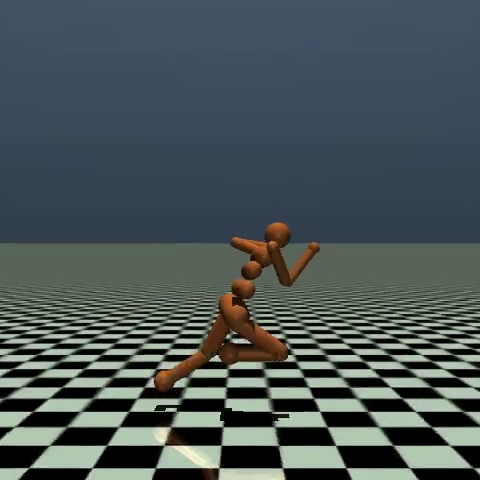}} 
\subfloat[step 72\label{subFig:dsac2}]
{\includegraphics[width = 0.2\textwidth, height = 0.24\textwidth]{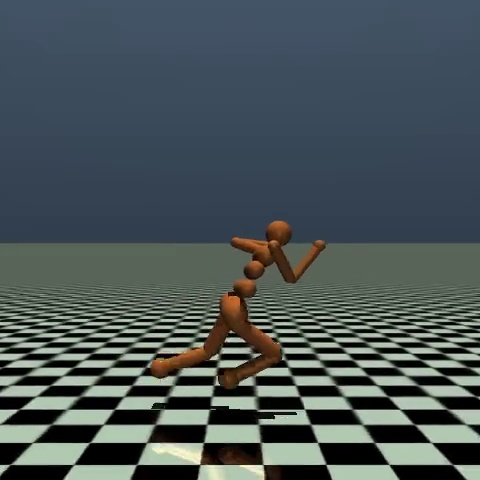}}
\subfloat[step 74\label{subFig:dsac3}]
{\includegraphics[width = 0.2\textwidth, height = 0.24\textwidth]{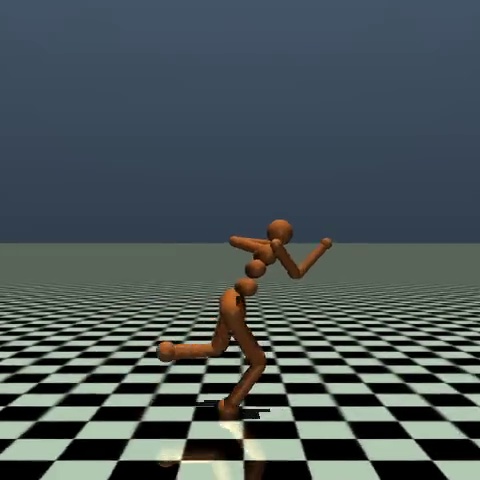}}
\subfloat[step 76\label{subFig:dsac4}]
{\includegraphics[width = 0.2\textwidth, height = 0.24\textwidth]{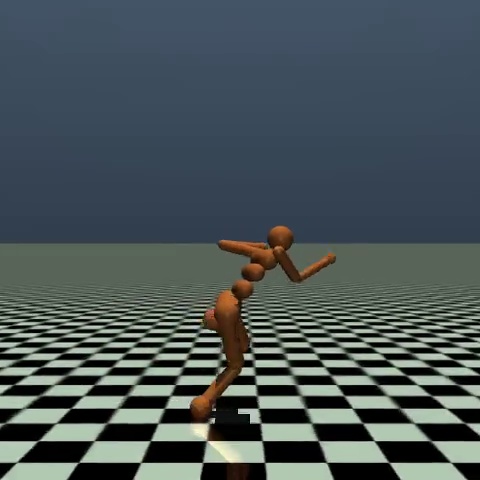}}
\subfloat[step 78\label{subFig:dsac5}]
{\includegraphics[width = 0.2\textwidth, height = 0.24\textwidth]{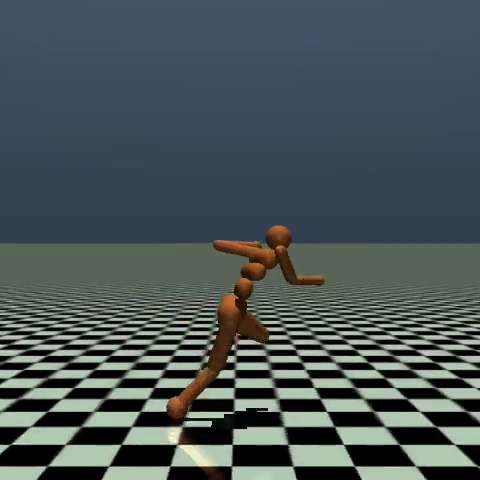}}\\
\vspace{0.8em}
\subfloat[\textbf{DSAC-T} step 70\label{subFig:dsact1}]
{\includegraphics[width = 0.2\textwidth, height = 0.24\textwidth]{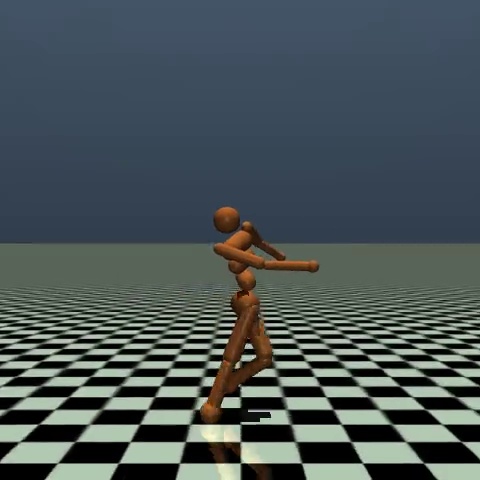}} 
\subfloat[step 72\label{subFig:dsact2}]
{\includegraphics[width = 0.2\textwidth, height = 0.24\textwidth]{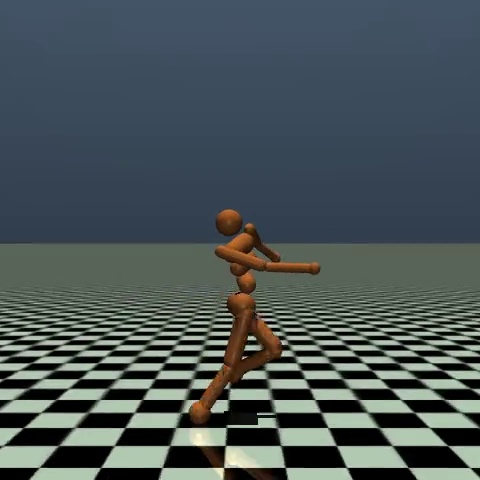}}
\subfloat[step 74\label{subFig:dsact3}]
{\includegraphics[width = 0.2\textwidth, height = 0.24\textwidth]{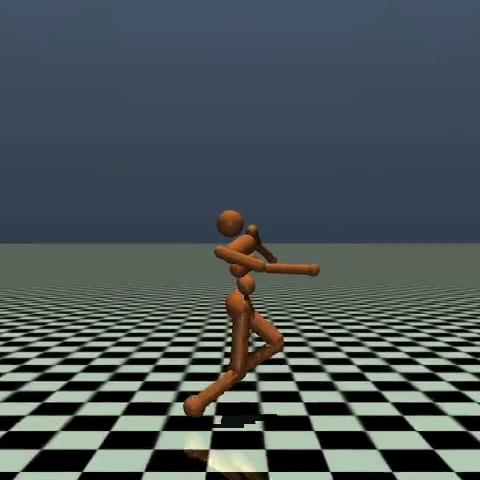}}
\subfloat[step 76\label{subFig:dsact4}]
{\includegraphics[width = 0.2\textwidth, height = 0.24\textwidth]{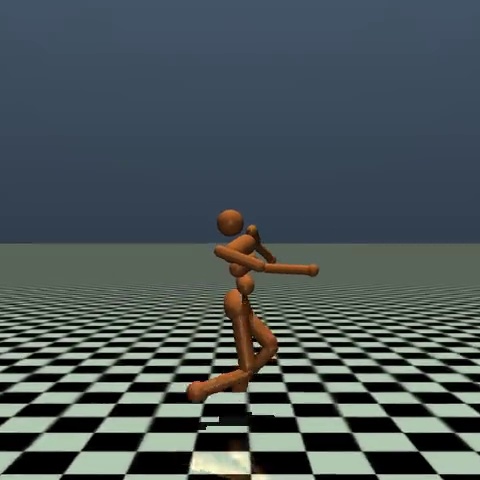}}
\subfloat[step 78\label{subFig:dsact5}]
{\includegraphics[width = 0.2\textwidth, height = 0.24\textwidth]{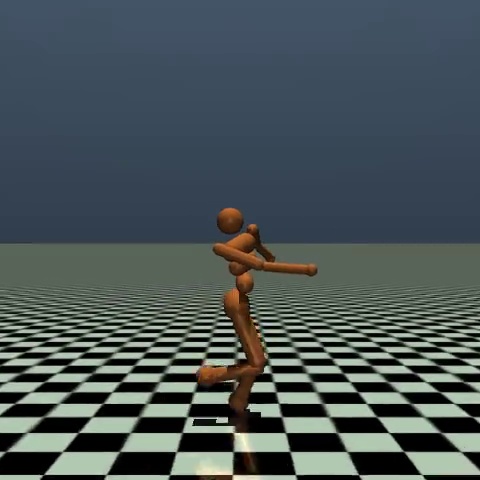}}\\
\vspace{0.8em}
\subfloat[\textbf{SAC} step 70\label{subFig:sac1}]
{\includegraphics[width = 0.2\textwidth, height = 0.24\textwidth]{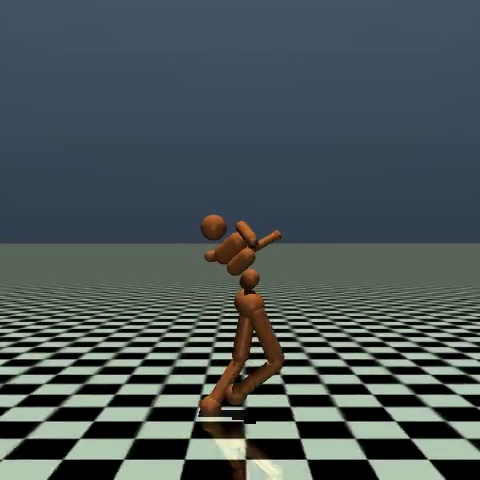}} 
\subfloat[step 72\label{subFig:sac2}]
{\includegraphics[width = 0.2\textwidth, height = 0.24\textwidth]{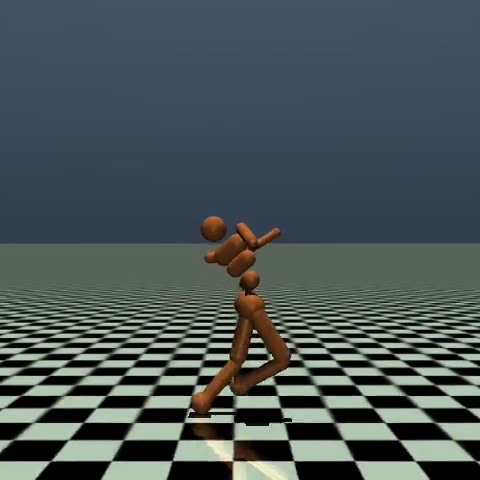}}
\subfloat[step 74\label{subFig:sac3}]
{\includegraphics[width = 0.2\textwidth, height = 0.24\textwidth]{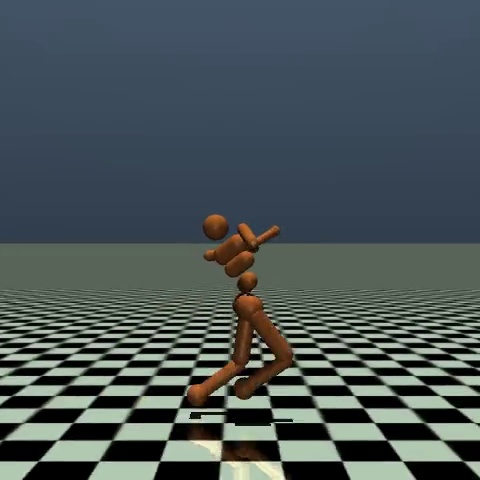}}
\subfloat[step 76\label{subFig:sac4}]
{\includegraphics[width = 0.2\textwidth, height = 0.24\textwidth]{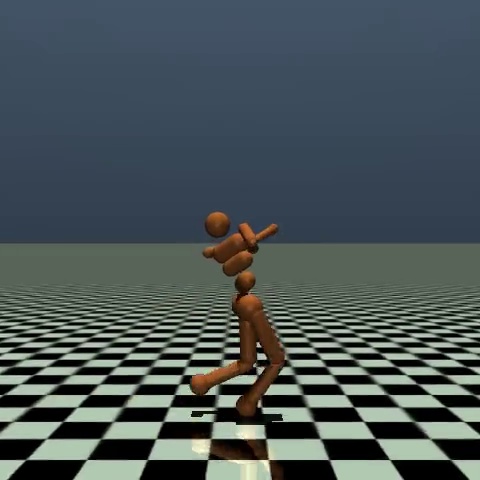}}
\subfloat[step 78\label{subFig:sac5}]
{\includegraphics[width = 0.2\textwidth, height = 0.24\textwidth]{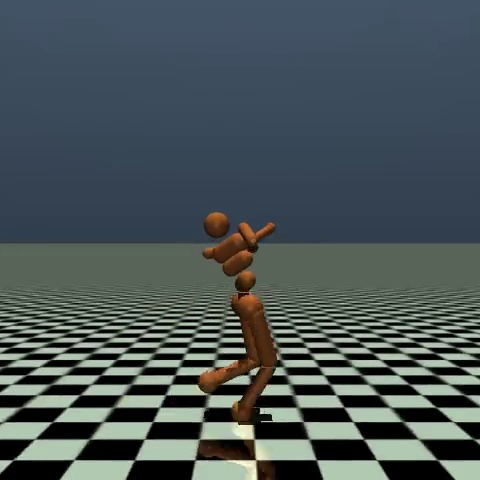}}\\

\caption{Visualizations of control processes on Humanoid-v3 task.
    }

\label{fig_video_frames}
\end{figure}

\newpage
\section{Reproducibility Statement}
\label{appen_hyper}




\begin{table}[h]
\centering
\captionsetup{justification=centering,labelsep=newline,font=small}
\captionsetup{justification=centering,labelsep=newline,font={small,sc}}
\caption{Detailed hyperparameters.}
\label{tab_hyper}
\begin{tabular}{lc}
\toprule
Hyperparameters & Value \\
\hline
\emph{Shared} & \\
\quad Optimizer &  Adam ($\beta_{1}=0.9, \beta_{2}=0.999$)\\
\quad Actor learning rate & $1{\rm{e-}}4 $\\
\quad Critic learning rate & $1{\rm{e-}}4 $\\
\quad Discount factor ($\gamma$) & 0.99 \\
\quad Policy update interval & 2\\
\quad Target smoothing coefficient ($\tau$) & 0.005\\
\quad Reward scale & 0.1\\
\quad Number of iterations & $1.5\times10^6$\\
\hline
\emph{Maximum-entropy framework} &\\ 
\quad  Learning rate of temperature $\alpha$ &  $3\times10^{-4}$ \\
\quad  Base expected entropy ($\overline{\mathcal{H}}$) &  $\overline{\mathcal{H}}=-{\rm{dim}}(\mathcal{A})$ \\
\hline
\emph{Deterministic policy} &\\ 
\quad Exploration noise&  $\epsilon \sim \mathcal{N}(0,0.1^2)$\\
\hline
\emph{Off-policy} &\\ 
\quad Sample batch size &  20 \\
\quad Replay batch size &  256 \\
\quad Replay buffer warm size & $1\times10^4$\\
\quad Replay buffer size & $1\times10^6$\\
\hline
\emph{On-policy} &\\ 
\quad Sample batch size &  2000 \\
\quad Replay batch size &  2000 \\
\quad GAE factor   & 0.95\\
\hline
\emph{DSAC-T} &\\ 
\quad Variance clipping constant $\zeta$  &  3 \\
\quad  Stabilizing constant $\epsilon$ and $\epsilon_{\omega}$  &  0.1 \\
\hline
\emph{DSAC-E (ours)} &\\ 
\quad $\rho$  &  20 for Humanoid and Walker2d, otherwise 1\\
\bottomrule
\end{tabular}
\end{table}


\paragraph{Time efficiency.}  The CPU used for the experiment is the AMD Ryzen
Threadripper 3960X 24-Core Processor, and the GPU is NVIDIA GeForce RTX 3090Ti. Taking
Humanoid-v3 as an example, the time taken to train 1.5 million iterations using the JAX framework around is 2 hours.

\newpage

\clearpage
\section{LLM Usage Disclosure}
We used ChatGPT to polish grammar and improve text clarity. We reviewed all LLM-generated suggestions and are fully responsible for the final content of this paper.

\end{document}

%% file: math_commands.tex
\usepackage{amsmath,amsfonts,bm}

\def\eqref#1{equation~\ref{#1}}

\def\1{\bm{1}}

\DeclareMathAlphabet{\mathsfit}{\encodingdefault}{\sfdefault}{m}{sl}
\SetMathAlphabet{\mathsfit}{bold}{\encodingdefault}{\sfdefault}{bx}{n}













%% file: main.bbl
\begin{thebibliography}{37}
\providecommand{\natexlab}[1]{#1}
\providecommand{\url}[1]{\texttt{#1}}
\expandafter\ifx\csname urlstyle\endcsname\relax
  \providecommand{\doi}[1]{doi: #1}\else
  \providecommand{\doi}{doi: \begingroup \urlstyle{rm}\Url}\fi

\bibitem[Amin et~al.(2021)Amin, Gomrokchi, Satija, Van~Hoof, and Precup]{amin2021survey}
Susan Amin, Maziar Gomrokchi, Harsh Satija, Herke Van~Hoof, and Doina Precup.
\newblock A survey of exploration methods in reinforcement learning.
\newblock \emph{arXiv preprint arXiv:2109.00157}, 2021.

\bibitem[An et~al.(2021)An, Moon, Kim, and Song]{an2021uncertainty}
Gaon An, Seungyong Moon, Jang-Hyun Kim, and Hyun~Oh Song.
\newblock Uncertainty-based offline reinforcement learning with diversified q-ensemble.
\newblock \emph{Advances in neural information processing systems}, 34:\penalty0 7436--7447, 2021.

\bibitem[Brockman et~al.(2016)Brockman, Cheung, Pettersson, Schneider, Schulman, Tang, and Zaremba]{brockman2016openaigym}
Greg Brockman, Vicki Cheung, Ludwig Pettersson, Jonas Schneider, John Schulman, Jie Tang, and Wojciech Zaremba.
\newblock Openai gym.
\newblock \emph{arXiv preprint arXiv:1606.01540}, 2016.

\bibitem[Celik et~al.(2025)Celik, Li, Blessing, Li, Palenicek, Peters, Chalvatzaki, and Neumann]{celik2025dime}
Onur Celik, Zechu Li, Denis Blessing, Ge~Li, Daniel Palenicek, Jan Peters, Georgia Chalvatzaki, and Gerhard Neumann.
\newblock Dime: Diffusion-based maximum entropy reinforcement learning.
\newblock \emph{arXiv preprint arXiv:2502.02316}, 2025.

\bibitem[Ding et~al.(2024)Ding, Hu, Zhang, Ren, Zhang, Yu, Wang, and Shi]{ding2024diffusion}
Shutong Ding, Ke~Hu, Zhenhao Zhang, Kan Ren, Weinan Zhang, Jingyi Yu, Jingya Wang, and Ye~Shi.
\newblock Diffusion-based reinforcement learning via q-weighted variational policy optimization.
\newblock \emph{Advances in Neural Information Processing Systems}, 37:\penalty0 53945--53968, 2024.

\bibitem[Duan et~al.(2021)Duan, Guan, Li, Ren, Sun, and Cheng]{duan2021distributional}
Jingliang Duan, Yang Guan, Shengbo~Eben Li, Yangang Ren, Qi~Sun, and Bo~Cheng.
\newblock Distributional soft actor-critic: Off-policy reinforcement learning for addressing value estimation errors.
\newblock \emph{IEEE transactions on neural networks and learning systems}, 33\penalty0 (11):\penalty0 6584--6598, 2021.

\bibitem[Duan et~al.(2025)Duan, Wang, Xiao, Gao, Li, Liu, Zhang, Cheng, and Li]{duan2025distributional}
Jingliang Duan, Wenxuan Wang, Liming Xiao, Jiaxin Gao, Shengbo~Eben Li, Chang Liu, Ya-Qin Zhang, Bo~Cheng, and Keqiang Li.
\newblock Distributional soft actor-critic with three refinements.
\newblock \emph{IEEE Transactions on Pattern Analysis and Machine Intelligence}, 2025.

\bibitem[Eysenbach \& Levine(2022)Eysenbach and Levine]{eysenbachmaximum}
Benjamin Eysenbach and Sergey Levine.
\newblock Maximum entropy rl (provably) solves some robust rl problems.
\newblock In \emph{International Conference on Learning Representations}, 2022.

\bibitem[Fox et~al.(2016)Fox, Pakman, and Tishby]{fox2016taming}
Roy Fox, Ari Pakman, and Naftali Tishby.
\newblock Taming the noise in reinforcement learning via soft updates.
\newblock In \emph{Proceedings of the Thirty-Second Conference on Uncertainty in Artificial Intelligence}, pp.\  202--211, 2016.

\bibitem[Fujimoto et~al.(2018)Fujimoto, van Hoof, and Meger]{Fujimoto2018TD3}
Scott Fujimoto, Herke van Hoof, and David Meger.
\newblock Addressing function approximation error in actor-critic methods.
\newblock In \emph{Proceedings of the 35th International Conference on Machine Learning (ICML 2018)}, pp.\  1587--1596, Stockholmsmässan, Stockholm Sweden, 2018. PMLR.

\bibitem[Fujimoto et~al.(2023)Fujimoto, Chang, Smith, Gu, Precup, and Meger]{fujimoto2023sale}
Scott Fujimoto, Wei-Di Chang, Edward Smith, Shixiang~Shane Gu, Doina Precup, and David Meger.
\newblock For sale: State-action representation learning for deep reinforcement learning.
\newblock \emph{Advances in neural information processing systems}, 36:\penalty0 61573--61624, 2023.

\bibitem[Haarnoja et~al.(2017)Haarnoja, Tang, Abbeel, and Levine]{haarnoja2017reinforcement}
Tuomas Haarnoja, Haoran Tang, Pieter Abbeel, and Sergey Levine.
\newblock Reinforcement learning with deep energy-based policies.
\newblock In \emph{International conference on machine learning}, pp.\  1352--1361. PMLR, 2017.

\bibitem[Haarnoja et~al.(2018{\natexlab{a}})Haarnoja, Zhou, Abbeel, and Levine]{Haarnoja2018SAC}
Tuomas Haarnoja, Aurick Zhou, Pieter Abbeel, and Sergey Levine.
\newblock Soft actor-critic: Off-policy maximum entropy deep reinforcement learning with a stochastic actor.
\newblock In \emph{Proceedings of the 35th International Conference on Machine Learning (ICML 2018)}, pp.\  1861--1870, Stockholmsmässan, Stockholm Sweden, 2018{\natexlab{a}}. PMLR.

\bibitem[Haarnoja et~al.(2018{\natexlab{b}})Haarnoja, Zhou, Hartikainen, Tucker, Ha, Tan, Kumar, Zhu, Gupta, Abbeel, et~al.]{Haarnoja2018ASAC}
Tuomas Haarnoja, Aurick Zhou, Kristian Hartikainen, George Tucker, Sehoon Ha, Jie Tan, Vikash Kumar, Henry Zhu, Abhishek Gupta, Pieter Abbeel, et~al.
\newblock Soft actor-critic algorithms and applications.
\newblock \emph{arXiv preprint arXiv:1812.05905}, 2018{\natexlab{b}}.

\bibitem[Hao et~al.(2023)Hao, Yang, Tang, Bai, Liu, Meng, Liu, and Wang]{hao2023exploration}
Jianye Hao, Tianpei Yang, Hongyao Tang, Chenjia Bai, Jinyi Liu, Zhaopeng Meng, Peng Liu, and Zhen Wang.
\newblock Exploration in deep reinforcement learning: From single-agent to multiagent domain.
\newblock \emph{IEEE Transactions on Neural Networks and Learning Systems}, 35\penalty0 (7):\penalty0 8762--8782, 2023.

\bibitem[Hazan et~al.(2019)Hazan, Kakade, Singh, and Van~Soest]{hazan2019provably}
Elad Hazan, Sham Kakade, Karan Singh, and Abby Van~Soest.
\newblock Provably efficient maximum entropy exploration.
\newblock In \emph{International Conference on Machine Learning}, pp.\  2681--2691. PMLR, 2019.

\bibitem[Hollenstein et~al.(2022)Hollenstein, Auddy, Saveriano, Renaudo, and Piater]{hollenstein2022action}
Jakob Hollenstein, Sayantan Auddy, Matteo Saveriano, Erwan Renaudo, and Justus Piater.
\newblock Action noise in off-policy deep reinforcement learning: Impact on exploration and performance.
\newblock \emph{arXiv preprint arXiv:2206.03787}, 2022.

\bibitem[Li et~al.(2021)Li, Huang, and Zhu]{li2021adaptive}
Min Li, Tianyi Huang, and William Zhu.
\newblock Adaptive exploration policy for exploration--exploitation tradeoff in continuous action control optimization.
\newblock \emph{International Journal of Machine Learning and Cybernetics}, 12\penalty0 (12):\penalty0 3491--3501, 2021.

\bibitem[Li(2023)]{li2023rlbook}
Shengbo~Eben Li.
\newblock \emph{Reinforcement Learning for Sequential Decision and Optimal Control}.
\newblock Springer Verlag, Singapore, 2023.

\bibitem[Lillicrap et~al.(2016)Lillicrap, Hunt, Pritzel, Heess, Erez, Tassa, Silver, and Wierstra]{lillDDPGicrap2015DDPG}
Timothy~P. Lillicrap, Jonathan~J. Hunt, Alexander Pritzel, Nicolas Heess, Tom Erez, Yuval Tassa, David Silver, and Daan Wierstra.
\newblock Continuous control with deep reinforcement learning.
\newblock In \emph{4th International Conference on Learning Representations (ICLR 2016)}, San Juan, Puerto Rico, 2016.

\bibitem[Nachum et~al.(2017)Nachum, Norouzi, Xu, and Schuurmans]{nachum2017bridging}
Ofir Nachum, Mohammad Norouzi, Kelvin Xu, and Dale Schuurmans.
\newblock Bridging the gap between value and policy based reinforcement learning.
\newblock In \emph{30th Advances in Neural Information Processing Systems (NeurIPS 2017)}, pp.\  2775--2785, Long Beach, CA, USA, 2017.

\bibitem[Plappert et~al.(2018)Plappert, Houthooft, Dhariwal, Sidor, Chen, Chen, Asfour, Abbeel, and Andrychowicz]{plappert2018parameter}
Matthias Plappert, Rein Houthooft, Prafulla Dhariwal, Szymon Sidor, Richard~Y Chen, Xi~Chen, Tamim Asfour, Pieter Abbeel, and Marcin Andrychowicz.
\newblock Parameter space noise for exploration.
\newblock In \emph{International Conference on Learning Representations}, 2018.

\bibitem[Rawlik et~al.(2012)Rawlik, Toussaint, and Vijayakumar]{rawlik2012stochastic}
Konrad Rawlik, Marc Toussaint, and Sethu Vijayakumar.
\newblock On stochastic optimal control and reinforcement learning by approximate inference.
\newblock \emph{Proceedings of Robotics: Science and Systems VIII}, 2012.

\bibitem[Schulman et~al.(2015)Schulman, Levine, Abbeel, Jordan, and Moritz]{schulman2015TRPO}
John Schulman, Sergey Levine, Pieter Abbeel, Michael~I. Jordan, and Philipp Moritz.
\newblock Trust region policy optimization.
\newblock In \emph{Proceedings of the 32nd International Conference on Machine Learning, (ICML 2015)}, pp.\  1889--1897, Lille, France, 2015.

\bibitem[Schulman et~al.(2017{\natexlab{a}})Schulman, Chen, and Abbeel]{schulman2017PG_Soft-Q}
John Schulman, Xi~Chen, and Pieter Abbeel.
\newblock Equivalence between policy gradients and soft q-learning.
\newblock \emph{arXiv preprint arXiv:1704.06440}, 2017{\natexlab{a}}.

\bibitem[Schulman et~al.(2017{\natexlab{b}})Schulman, Wolski, Dhariwal, Radford, and Klimov]{schulman2017PPO}
John Schulman, Filip Wolski, Prafulla Dhariwal, Alec Radford, and Oleg Klimov.
\newblock Proximal policy optimization algorithms.
\newblock \emph{arXiv preprint arXiv:1707.06347}, 2017{\natexlab{b}}.

\bibitem[Seo et~al.(2025)Seo, Sferrazza, Geng, Nauman, Yin, and Abbeel]{seo2025fasttd3}
Younggyo Seo, Carmelo Sferrazza, Haoran Geng, Michal Nauman, Zhao-Heng Yin, and Pieter Abbeel.
\newblock Fasttd3: Simple, fast, and capable reinforcement learning for humanoid control.
\newblock \emph{arXiv preprint arXiv:2505.22642}, 2025.

\bibitem[Sikchi et~al.(2022)Sikchi, Zhou, and Held]{sikchi2022learning}
Harshit Sikchi, Wenxuan Zhou, and David Held.
\newblock Learning off-policy with online planning.
\newblock In \emph{Conference on Robot Learning}, pp.\  1622--1633. PMLR, 2022.

\bibitem[Sun et~al.(2022)Sun, Han, Yang, Ma, Guo, and Zhou]{sun2022exploit}
Hao Sun, Lei Han, Rui Yang, Xiaoteng Ma, Jian Guo, and Bolei Zhou.
\newblock Exploit reward shifting in value-based deep-rl: Optimistic curiosity-based exploration and conservative exploitation via linear reward shaping.
\newblock \emph{Advances in neural information processing systems}, 35:\penalty0 37719--37734, 2022.

\bibitem[Sutton \& Barto(2018)Sutton and Barto]{sutton2018reinforcement}
Richard~S Sutton and Andrew~G Barto.
\newblock \emph{Reinforcement learning: {A}n introduction}.
\newblock MIT press, 2018.

\bibitem[Tokic(2010)]{tokic2010adaptive}
Michel Tokic.
\newblock Adaptive $\varepsilon$-greedy exploration in reinforcement learning based on value differences.
\newblock In \emph{Annual conference on artificial intelligence}, pp.\  203--210. Springer, 2010.

\bibitem[Wang et~al.(2022)Wang, Wang, Liang, Zhao, Huang, Xu, Dai, and Miao]{wang2022deep}
Xu~Wang, Sen Wang, Xingxing Liang, Dawei Zhao, Jincai Huang, Xin Xu, Bin Dai, and Qiguang Miao.
\newblock Deep reinforcement learning: A survey.
\newblock \emph{IEEE Transactions on Neural Networks and Learning Systems}, 35\penalty0 (4):\penalty0 5064--5078, 2022.

\bibitem[Wang et~al.(2024)Wang, Wang, Jiang, Zou, Liu, Song, Wang, Xiao, Wu, Duan, et~al.]{wang2024diffusion}
Yinuo Wang, Likun Wang, Yuxuan Jiang, Wenjun Zou, Tong Liu, Xujie Song, Wenxuan Wang, Liming Xiao, Jiang Wu, Jingliang Duan, et~al.
\newblock Diffusion actor-critic with entropy regulator.
\newblock \emph{Advances in Neural Information Processing Systems}, 37:\penalty0 54183--54204, 2024.

\bibitem[Wang et~al.(2025)Wang, Tan, Zou, Lin, Song, Wang, Liu, Wang, Zhan, Zhu, et~al.]{wang2025enhanced}
Yinuo Wang, Mining Tan, Wenjun Zou, Haotian Lin, Xujie Song, Wenxuan Wang, Tong Liu, Likun Wang, Guojian Zhan, Tianze Zhu, et~al.
\newblock Enhanced dacer algorithm with high diffusion efficiency.
\newblock \emph{arXiv preprint arXiv:2505.23426}, 2025.

\bibitem[Yang et~al.(2023{\natexlab{a}})Yang, Zhang, Song, Hong, Xu, Zhao, Zhang, Cui, and Yang]{yang2023diffusion}
Ling Yang, Zhilong Zhang, Yang Song, Shenda Hong, Runsheng Xu, Yue Zhao, Wentao Zhang, Bin Cui, and Ming-Hsuan Yang.
\newblock Diffusion models: A comprehensive survey of methods and applications.
\newblock \emph{ACM computing surveys}, 56\penalty0 (4):\penalty0 1--39, 2023{\natexlab{a}}.

\bibitem[Yang et~al.(2023{\natexlab{b}})Yang, Huang, Lei, Zhong, Yang, Fang, Wen, Zhou, and Lin]{yang2023policy}
Long Yang, Zhixiong Huang, Fenghao Lei, Yucun Zhong, Yiming Yang, Cong Fang, Shiting Wen, Binbin Zhou, and Zhouchen Lin.
\newblock Policy representation via diffusion probability model for reinforcement learning.
\newblock \emph{arXiv preprint arXiv:2305.13122}, 2023{\natexlab{b}}.

\bibitem[Zhu et~al.(2023)Zhu, Zhao, He, Zhong, Zhang, Guo, Chen, and Zhang]{zhu2023diffusion}
Zhengbang Zhu, Hanye Zhao, Haoran He, Yichao Zhong, Shenyu Zhang, Haoquan Guo, Tingting Chen, and Weinan Zhang.
\newblock Diffusion models for reinforcement learning: A survey.
\newblock \emph{arXiv preprint arXiv:2311.01223}, 2023.

\end{thebibliography}
